%% file: WMCar_preprint.tex
\documentclass[10pt,journal,compsoc]{IEEEtran}

\usepackage{soul}

\usepackage{times}
\usepackage{epsfig}
\usepackage{graphicx}
\usepackage{amsmath}
\usepackage{amssymb}

\usepackage{mathrsfs}
\usepackage{color}
\usepackage{epstopdf}
\usepackage{multirow}
\usepackage{makebox}

\usepackage{booktabs}

\usepackage{mathtools}
\DeclarePairedDelimiter\ceil{\lceil}{\rceil}

\usepackage{caption}
\usepackage{subcaption}

\usepackage{graphics}
\usepackage{algorithm}
\usepackage{algpseudocode}
\algnewcommand{\LineComment}[1]{\State \(\triangleright\) #1}
\newcommand{\head}[1]{\textnormal{\textbf{#1}}}

\input{short.tex}

\graphicspath{{fig_MCar/}}

\ifCLASSOPTIONcompsoc
  \usepackage[nocompress]{cite}
\else
  \usepackage{cite}
\fi

\ifCLASSINFOpdf
\else
\fi

\begin{document}
%\title{MCar with ICE}
%\title{Learning from Imbalanced \\ Ambiguously Labeled Data}

\title{Learning from Ambiguously Labeled \\ Face Images}

\author{Ching-Hui Chen,
        Vishal M. Patel,~\IEEEmembership{Senior Member,~IEEE,}
        Rama Chellappa,~\IEEEmembership{Fellow,~IEEE}% <-this % stops a space

\IEEEcompsocitemizethanks{\IEEEcompsocthanksitem C.-H. Chen and R. Chellappa are with the Department of Electrical and
Computer Engineering, University of Maryland, College Park, MD,
20742 USA (e-mail:\{ching, rama\}@umiacs.umd.edu).
\IEEEcompsocthanksitem V. M. Patel is with the Department of Electrical and Computer
Engineering, Rutgers University, Piscataway, NJ, 08901 USA (e-mail:
vishal.m.patel@rutgers.edu).
}% <-this % stops an unwanted space
}

\IEEEtitleabstractindextext{%
\begin{abstract}
Learning a classifier from ambiguously labeled face images is challenging since training images are not always explicitly-labeled. For instance, face images of two persons in a news photo are not explicitly labeled by their names in the caption. We propose a Matrix Completion for Ambiguity Resolution (MCar) method for predicting the actual labels from ambiguously labeled images. This step is followed by learning a standard supervised classifier from the disambiguated labels to classify new images. To prevent the majority labels from dominating the result of MCar, we generalize MCar to a weighted MCar (WMCar) that handles label imbalance. Since WMCar outputs a soft labeling vector of reduced ambiguity for each instance, we can iteratively refine it by feeding it as the input to WMCar. Nevertheless, such an iterative implementation can be affected by the noisy soft labeling vectors, and thus the performance may degrade. Our proposed Iterative Candidate Elimination (ICE) procedure makes the iterative ambiguity resolution possible by gradually eliminating a portion of least likely candidates in ambiguously labeled face. We further extend MCar to incorporate the labeling constraints between instances when such prior knowledge is available. Compared to existing methods, our approach demonstrates improvement on several ambiguously labeled datasets.
\end{abstract}

\begin{IEEEkeywords}
Ambiguous learning, labeling imbalance, iterative candidate elimination, matrix completion, low-rank matrix recovery.
\end{IEEEkeywords}}

\maketitle

\IEEEraisesectionheading{\section{Introduction}\label{sec:introduction}}

\begin{figure}[t]
\centering
\includegraphics[trim= 195 130 170 160,clip,width=0.4\textwidth]{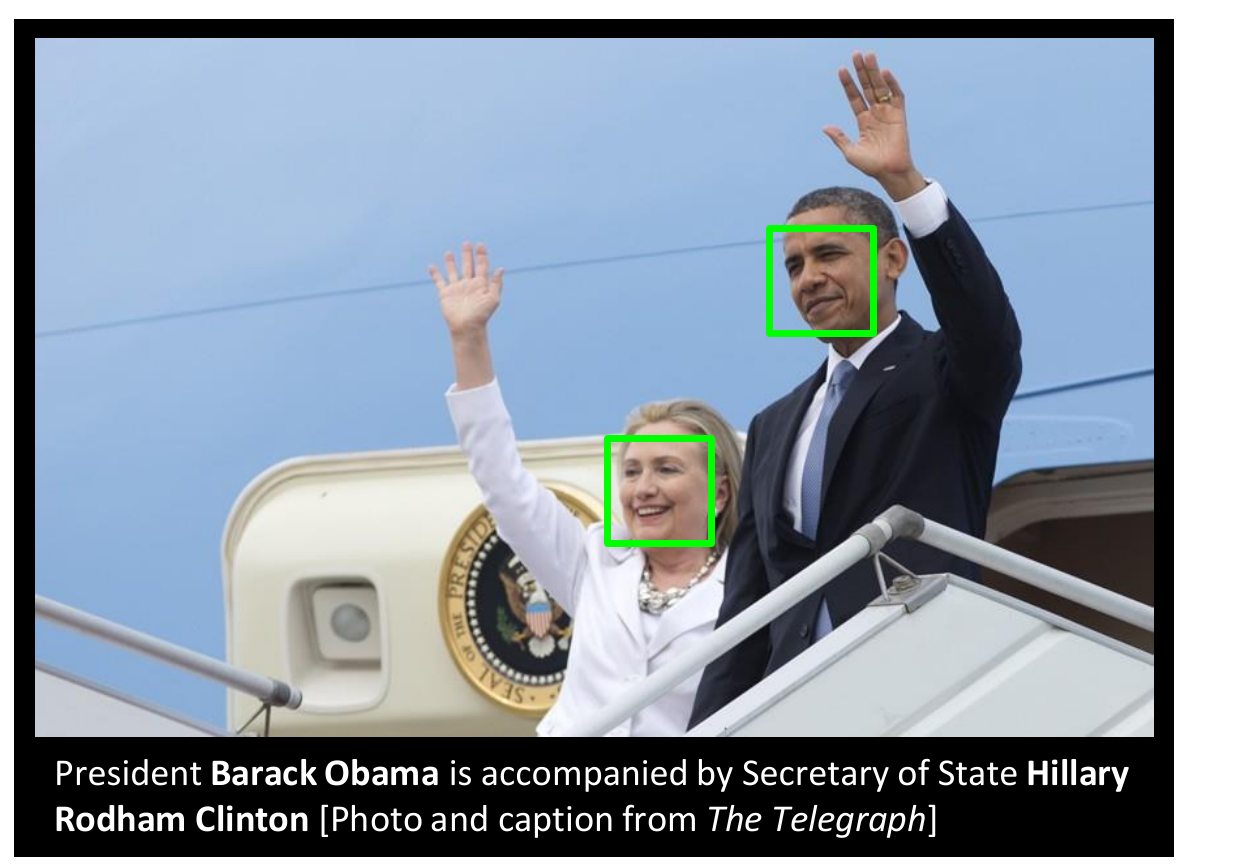}
\caption{The names in the captions are not explicitly associated with the face images in the news photo.}
\label{fig:newsphoto}
\end{figure}

\IEEEPARstart{L}{earning} a classifier for naming a face requires a large amount of labeled face images and videos.  However, labeling face images is expensive and time-consuming due to significant amount of human efforts involved.  As a result, brief descriptions such as tags, captions and screenplays accompanying the images and videos become important for training the classifiers.  Although such information is publicly available, it is not as explicitly labeled as human annotations.  For instance, names in the caption of a news photo provide possible candidates for faces appearing in the image \cite{Berg2004,Berg20041} (see Figure~\ref{fig:newsphoto}).  The names in the screenplays are only weakly associated with faces in the shots \cite{Everingham2006ReID}.  The problem in which instead of a single label per instance, one is given a candidate set of labels, of which only one is correct is known as ambiguously labeled learning\footnote{also known as partially labeled learning and superset label learning} \cite{Hullermeier2006,Cour2009,Liu2012acm,Chen2013,Liu2014lot}.

In recent years, the problem of completing a low-rank matrix with missing entries has gained significant attention. In particular, matrix completion methods have been shown to produce good results for multi-label image classification problems \cite{Goldberg2010}, \cite{Cabral2011}. In these methods, the underlying assumption is that the concatenation of feature vectors and their labels produce a low-rank matrix.  Our work is motivated by these works.  The proposed method, Matrix Completion for Ambiguity Resolution (MCar), takes the heterogeneous feature matrix, which is the concatenation of the labeling matrix and feature matrix, as input. We first show that the heterogeneous feature matrix is ideally low-rank in the absence of noise.  This in turn, allows us to convert the labeling problem as a matrix completion problem by pursuing the underlying low-rank matrix of the heterogeneous feature matrix. In contrast to multi-label learning, ambiguous labeling provides the clue that one of the labels in the candidate label set is the true label. This knowledge is utilized to regularize the labeling matrix in the heterogeneous feature matrix.  This is essentially the main difference between our work and some of the previously proposed matrix completion techniques \cite{Goldberg2010}, \cite{Cabral2011}.

Although ambiguous learning techniques can take advantage of large-scale and diverse ambiguously labeled data, most methods cannot properly handle the labeling imbalance that is often present in publicly available training data. For instances, celebrities and leading actors usually dominate (appear more frequently) in the candidate label sets, and these majority labels can easily bias the results of ambiguity resolving methods.  As the proposed method relies on low-rank approximation of the heterogeneous feature matrix, heterogeneous feature vectors associated with those majority labels can dominate the process of low-rank approximation and thus bias the recovery of the labeling matrix. We propose the weighted MCar (WMCar) to overcome the labeling imbalance in ambiguously labeled data. Unlike conventional instance weighting techniques \cite{He2009lfi} that assign unequal instance weight to the cost function of instances, WMCar performs unequal column-wise weighting on the heterogeneous feature vectors. Therefore, a heterogeneous feature vector associated with majority labels will contribute less to the process of low-rank approximation than that associated with minority labels.

The column-wise weighting in WMCar can be computed by estimating the groundtruth label distribution from the recovered labeling matrix, but the recovered labeling matrix is not accessible without applying WMCar to resolve the ambiguity in the original labeling matrix.
Nevertheless, iteratively updating the column-wise weighting and recovering the labeling matrix with WMCar is not reliable (see iterative WMCar in Figure \ref{fig:WMCar_ICE_thre}). An explanation is that there is some unresolved ambiguity in the soft labeling matrix recovered by WMCar. The remaining ambiguity (noise) can be detrimental to the iterative process as we iteratively update WMCar by substituting the labeling matrix with the recovered one from the previous iteration.
Hence, we propose the Iterative Candidate Elimination (ICE) procedure to iteratively eliminate the least likely candidates from a portion of the ambiguously labeled data.
This procedure iteratively suppresses the noise in the recovered labeling matrix and thus yields a better performance in the next iteration of WMCar. Although WMCar with ICE is an iterative approach, it is fundamentally different from previously suggested iterative methods \cite{Zeng2013,Chen2013,Chen2014all}. Unlike previous works that iteratively construct class-specific models and update the labels, the iterative process of ICE is effective in sequential noise suppression. Besides, WMCar concatenates the labels and features as a heterogeneous matrix to recover the labels in each iteration. This ensures that the information in the ambiguously labeled data is used as a whole in recovering the true labels.

Moreover, we generalize MCar to include the labeling constraints between the instances for practical applications. For example, two persons in a news photo should not be identified as the same subject even though both of them are ambiguously labeled in the caption.
As shown by the recent success in low-rank matrix recovery \cite{Candes2011}, several prior works have developed robust methods for classification  \cite{Chen2012}, \cite{Huang2012}. The proposed method inherits the benefit of low-rank matrix recovery and possesses the capability to resolve the label ambiguity via low-rank approximation of the heterogeneous matrix.  As a result, our method is more robust compared to some of the existing discriminative ambiguous learning methods \cite{Cour2009,Luo2010}.  The disambiguated labels from MCar are then used to learn a supervised learning classifier.

In this paper, we make the following contributions:\\
\noindent {\bf{1.}} We propose a matrix completion method where instances and their associated ambiguous labels are jointly considered for disambiguating the class labels.\\
\noindent {\bf{2.}} We provide a geometric interpretation of the matrix completion framework from the perspective of recovering the potentially-separable convex hulls of each class. \\
\noindent {\bf{3.}}  We propose WMCar to resolve the label ambiguity in the presence of labeling imbalance.\\
\noindent {\bf{4.}} We propose the ICE approach to improve the reliability of iterative WMCar. The integration of WMCar and ICE is effective in resolving the ambiguity and outperforms WMCar in general. \\
\noindent {\bf{5.}} Our method can handle the group constraints between instances for practical applications.

In this paper, we generalize our prior work in \cite{Chen2015mcf} to overcome labeling imbalance in ambiguously labeled data. The ICE procedure and the experimental analysis are extensions to \cite{Chen2015mcf}.

The rest of this paper is organized as follows. In Section
\ref{Chapter_MCar:sec:Related} we review some related work on ambiguously labeled learning methods.  Section \ref{Chapter_MCar:sec:Approach} describes the proposed MCar and WMCar. The optimization procedure for WMCar is described in Section \ref{MCar_sec:opt}. Section \ref{sec:ice} describes the ICE procedure in detail.  Section \ref{sec:lpc} presents the extension of MCar for incorporating the constraint between instances. In Section \ref{Chapter_MCar:sec:results}, we demonstrate the results on synthesized as well as real-world ambiguously labeled datasets. Finally, Section \ref{Chapter_MCar:sec:Con} concludes this work with a brief summary and discussion.

{\bf Notations:}
We use the following notations in this paper.
The matrix element $a_{i,j}$ denotes the entity in the $i^{\mathrm{th}}$ row and $j^{\mathrm{th}}$ column of matrix $\mathbf{A}$.
$\mathbf{1}_{n}$ represents a column vector of size $n \times 1$ consisting of 1's as its entries. $\mathbf{v}_{i}$ is the canonical vector corresponding to the 1-of-K coding of $i$.
$\|\cdot\|_1$ and $\|\cdot\|_{0}$ denote the $\ell_{1}$ norm and $\ell_{0}$ norm, respectively.
The Frobenius norm and the nuclear norm of  $\mathbf{A}$  are defined as $\|\mathbf{A}\|_F=\left(\sum_{i,j}(a_{i,j})^{2}\right)^\frac{1}{2}$ and $\|\mathbf{A}\|_{*}=\sum_{i}\sigma_{i}(\mathbf{A})$, respectively where $\sigma_{i}$ is the $i^{th}$ singular value of $\mathbf{A}$.
$(\cdot)^T$ denotes transposition operation. $|S|$ returns the cardinality in set $S$. $\mathcal{S}_a[b] = \mathrm{sgn}(b) \max(|b|-a,0)$ is the shrinkage operator. The concatenation of matrix $\mathbf{A}$ and $\mathbf{B}$ is defined as $\begin{bmatrix}
\mathbf{A} \\
\mathbf{B}
\end{bmatrix} = [\mathbf{A} ; \mathbf{B}]$.

\section{Related Work} \label{Chapter_MCar:sec:Related}
Various methods have been proposed in the literature for dealing with ambiguously labeled data.  Some of these methods propose Expectation Maximization (EM)-like approaches to alternately disambiguate the labels and learn a discriminative classifier \cite{Ambroise2001,Jin2002}. Berg \emph{et al.} \cite{Berg2004} proposed an EM-like approach to alternately disambiguate the labels by maximizing the likelihood of label assignment and estimate the parameters for the appearance model and language model. Non-parametric methods have also been used to resolve the ambiguity by leveraging the inductive bias of learning methods \cite{Hullermeier2006}.  For the ambiguously labeled training data the actual loss of mislabeling is not explicit.   As a result, it is difficult to learn an effective discriminative model.  Cour \emph{et al.} \cite{Cour2009,Cour2011} proposed the partial 0/1 loss function for ambiguous labeling, which is a tighter upper bound for the actual loss as compared to the 0/1 loss \cite{Zhang2004}. Subsequently, a discriminative classifier can be learned from the ambiguous labels by minimizing the partial 0/1 loss. Several works have improved the learning of partial labels with the modeling of partial loss \cite{Cid-sueiro2012plf}, error-correcting output codes \cite{Zhang2014dfp}, and iterative label propagation \cite{Zhang2015stp}. Liu \emph{et al.} \cite{Liu2012acm} proposed to learn a conditional multinomial mixture model for predicting the actual label from ambiguous labels.

Several dictionary-based methods have also been proposed for handing partially labeled datasets \cite{Ashish_PR2015,Chen2013,Chen2014all}.  In particular, an EM-like dictionary learning approach was proposed in \cite{Chen2013}, where a confidence matrix and dictionary are updated in alternating iterations. Although several methods have been utilizing the EM-like framework with robust appearance models \cite{Berg2004, Zeng2013,Ashish_PR2015,Chen2013,Chen2014all}, these methods can be very sensitive to the initialization of the model and may suffer from suboptimal performance. On the other hand, our proposed framework unifies the ambiguity resolution and appearance modeling into a single matrix completion framework, and thus it is more effective in ambiguity resolution.

Luo \emph{et al.} \cite{Luo2010} generalize the ambiguously labeled learning problem addressed in \cite{Cour2009} from single instances to group instances. The ambiguous loss considers the association between the group of identities and the candidate label vectors. The pairwise constraint between the instances (e.g. unique appearance of a subject) is accounted for when generating the candidate label vectors.  Furthermore, Zeng \emph{et al.} \cite{Zeng2013} use a Partial Permutation Matrix (PPM) to associate the identities in a group with ambiguous labels.  The pairwise constraint is encoded by restricting the structure of PPM.  Assuming that instances of the same subject inferred by PPM can ideally form a low-rank matrix, the actual identity of an instance can be predicted by alternatively updating the low-rank subspace and PPM. Xiao \emph{et al.} \cite{Xiao2015afn} associate the identities in a group from ambiguous labels by minimizing the summation of the discriminative affinities in a group, where the affinities are learned from the low-rank reconstruction coefficient matrix and the weak supervision of ambiguous labels.

Recently, learning from weak annotations of labeling imbalance has received significant attention \cite{Sahare2012aro,Zhang2015tci}. Chen \emph{et al.} \cite{Chen2006eco} employ the part-versus-part decomposition \cite{Lu2004apv} to overcome the data imbalance in multi-label learning. Charte \emph{et al.} \cite{Charte2015aii} propose several methods to resample the multi-label training data to compensate the imbalance level. Wu \emph{et al.} \cite{Wu2016csm} incorporate the class cardinality bound constraints to deal with class imbalance. Although several prior works have addressed the issue of imbalanced data in the context of multi-label learning, the labeling imbalance in ambiguously labeled data remains to be investigated.
We propose to estimate the groundtruth label distribution from ambiguous labels. With the estimated groundtruth label distribution, the instance weight of WMCar can be computed to deal with labeling imbalance.

\section{The Proposed Framework} \label{Chapter_MCar:sec:Approach}
The ambiguously labeled data is denoted as $\mathcal{L} = \{(\mathbf{x}_j, L_j),\,\, j=1, 2, \dots, N\}$, where $N$ is the number of instances. There are $c$ classes, and the class labels are denoted as $\mathcal{Y} =\{1, 2, \dots, c\}$. Note that $\mathbf{x}_j$ is the feature vector of the $j^{th}$ instance, and its candidate labeling set $L_j \subseteq \mathcal{Y}$ consists of candidate labels associated with the $j^{th}$ instance.
The true label of the $j^{th}$ instance is $l_j \in L_j$. In other words, one of the labels in $L_j$ is the true label of $\mathbf{x}_j$. The objective is to resolve the ambiguity in $\mathcal{L}$ such that each predicted label $\hat{l}_j$ of $\mathbf{x}_j$ matches its true label $l_j$.
We associate the candidate labeling set $L_j$ with a soft labeling vector $\mathbf{p}_j$, where $p_{i,j}$ indicates the probability that instance $j$ belongs to class $i$. This allows us to quantitatively assign the likelihood of each class the instance belongs to if such information is provided.
Given the ambiguous label of the $j^{th}$ instance, we assign each entry of $\mathbf{p}_j$ as
\begin{equation}
\begin{aligned}
\left\{
\begin{array}{ll}
p_{i,j} \in (0, 1]  \quad\,\,\, \mathrm{if}  \,\, i \in L_j,\, \\
p_{i,j} = 0 \quad\,\quad\quad \mathrm{if} \,\, i \notin L_j,\,
 \end{array} \right.
\end{aligned} j = 1, 2, \dots, N,
\end{equation}
where $\sum_{i=1}^c p_{i,j} = 1$. Without any prior knowledge, we assume equal probability for each candidate label.  Let $\mathbf{P} \in \mathbb{R}^{c \times N}$ denote the ambiguous labeling matrix with $\mathbf{p}_j$ in its $j^{th}$ column.  With this, one can model the ambiguous labeling as
\begin{equation}
\begin{aligned}
\mathbf{P}^0 = \mathbf{P} - \mathbf{E}_P, \label{eq:P}
\end{aligned}
\end{equation}
where $\mathbf{P}^0$ and $\mathbf{E}_P$ denote the true labeling matrix and the labeling noise, respectively.  The $j^{th}$ column vector of $\mathbf{P}^0$ is $\mathbf{p}^0_j = \mathbf{v}_{l_j}$, where $\mathbf{v}_{l_j}$ is the canonical vector corresponding to the 1-of-K coding of its true label $l_j$.

Similarly, assuming that the feature vectors are corrupted by some noise or occlusion, the feature matrix $\mathbf{X}$ with $\mathbf{x}_j$ in its $j^{th}$ column can be modeled as
\begin{equation}
\begin{aligned}
\mathbf{X}^0  = \mathbf{X} - \mathbf{E}_X, \label{eq:X}
\end{aligned}
\end{equation}
\textcolor[rgb]{0.00,0.00,0.00}{where $\mathbf{X} \in \mathbb{R}^{m \times N}$ consists of $N$ feature vectors of dimension $m$, $\mathbf{X}^0$ represents the feature matrix in the absence of noise and $\mathbf{E}_X$ accounts for the noise. }  Concatenating (\ref{eq:P}) and (\ref{eq:X}), we obtain a unified model of ambiguous labels and feature vectors, which can be expressed as
\begin{equation}
\begin{aligned}
\begin{bmatrix}
\mathbf{P}^0 \\
\mathbf{X}^0
\end{bmatrix}
=
\begin{bmatrix}
\mathbf{P} \\
\mathbf{X}
\end{bmatrix}
-
\begin{bmatrix}
\mathbf{E}_P \\
\mathbf{E}_X
\end{bmatrix}.
\end{aligned}\label{eq:united_model}
\end{equation}
Let
\begin{equation}
\begin{aligned}
\mathbf{H}_{obs} = \begin{bmatrix}
\mathbf{P} \\
\mathbf{X}
\end{bmatrix} \, \mathrm{and} \,\, \mathbf{E} = \begin{bmatrix}
\mathbf{E}_P \\
\mathbf{E}_X
\end{bmatrix}
\end{aligned}
\end{equation}
denote the heterogeneous feature matrix and its noise, respectively.  If we can show that $\mathbf{H}_{obs}$ is a low-rank matrix in the absence of noise, then we can use matrix completion methods for resolving the ambiguity in labeling.  In the following section, we investigate the low-rank property of $\mathbf{H}_{obs}.$

\subsection{Exploiting the Rank of $\mathbf{H}_{obs}$} \label{subsec:lowrank_obs}
The column vectors of $\mathbf{X}_0$ can be partitioned into sets $S_1, S_2, \dots, S_c$ based on their true labels. We assume that the elements of $S_k$ form a convex hull $C_k$ of $n_k$ vertices.  It is clear that $n_k \leq |S_k|$.
The representative matrix of the $k^{th}$class, $\mathbf{D}_k \in \mathbb{R}^{m \times n_k}$, consists of vertices of $C_k$ as its column vectors, and each column vector is treated as a representative of the $k^{th}$class.
Therefore, according to the definition of a convex hull, a noise-free instance $\mathbf{x}_j^0$ from class $k$ ($\mathbf{x}_j^0 \in C_k$) can be represented as
\begin{equation}
\begin{aligned}
\mathbf{x}^{0}_j = \mathbf{D}_k \mathbf{a}_{k,j}, \,\, \mathrm{where} \,\, \mathbf{a}_{k,j}^T \mathbf{1}_{n_k} =1, \mathbf{a}_{k,j} \in \mathbb{R}_+^{n_k \times 1}. \label{eq:convexhull}
\end{aligned}
\end{equation}
Note that $\mathbf{a}_{k,j} \in \mathbb{R}_+^{n_k \times 1}$ is the coefficient vector associated with the representative matrix of the $k^{th}$ class.
As the true label of an instance is not known in advance, we can represent $\mathbf{x}^{0}_j$ as
\begin{equation}
\begin{aligned}
\mathbf{x}^{0}_j &= \mathbf{D}\mathbf{q}_j, \quad
\mathbf{D} = [\mathbf{D}_1 \,\, \mathbf{D}_2 \,\, \cdots \,\, \mathbf{D}_c], \\
\mathbf{q}_j &= [\mathbf{a}_{1,j}^T \,\, \mathbf{a}_{2,j}^T \,\, \cdots \,\, \mathbf{a}_{c,j}^T]^T, \,\, \mathbf{q}_j^T \mathbf{1} =1, \label{eq:x_j}
\end{aligned}
\end{equation}
where $\mathbf{D} \in \mathbb{R}^{m \times (\sum_{i=1}^c n_i)}$ is the collective representative matrix, and $\mathbf{q}_j \in \mathbb{R}_+^{(\sum_{i=1}^c n_i) \times 1}$ is the associated coefficient vector.

According to (\ref{eq:x_j}), we can decompose $\mathbf{X}^0$ as
\begin{equation}
\begin{aligned}
\mathbf{X}^0 = \mathbf{D}\mathbf{Q}. \label{eq:X0_decomp}
\end{aligned}
\end{equation}
The coefficient matrix $\mathbf{Q}$ in (\ref{eq:X0_decomp}) is not unique as column vectors of $\mathbf{D}$ are not necessarily linearly independent.
However, we assume that an ideal decomposition $\mathbf{X}^0 =  \mathbf{D}\mathbf{Q}^* $ satisfies the following condition
\begin{equation}
\begin{aligned}
\mathbf{x}^{0}_j =  \mathbf{D}  \mathbf{q}^*_j,   \,\, \mathrm{where} \,\, &\mathbf{a}_{k,j}^{*T} \mathbf{1}_{n_k} = 1,\,\, \mathbf{x}_j^0 \in S_k,     \\
&\mathbf{a}_{l,j}^{*T} \mathbf{1}_{n_l} = 0,\,\, l \neq k,
\label{eq:convexhull_cat}
\end{aligned}
\end{equation}
which implies that $\mathbf{x}_j^0$ is exclusively represented by $\mathbf{D}_k$ even though it is possible that it can be written as a linear combination of any other vertices from different classes.

With this, we can recover the true labels from
\begin{equation}
\begin{aligned}
\mathbf{P}^0 = \mathbf{T}\mathbf{Q}^*,
\label{eq:P_recover}
\end{aligned}
\end{equation}
where $\mathbf{T} = [\mathbf{v}_1\mathbf{1}^T_{n_1} \,\, \mathbf{v}_2 \mathbf{1}^T_{n_2} \,\, \cdots \,\, \mathbf{v}_c \mathbf{1}^T_{n_c} ]$ accumulates the coefficients associated with each matrix representative.
Hence, the coefficient vector of dimension $\sum_{i=1}^c n_i$ is converted into labeling vector of dimension $c$. Concatenating $\mathbf{P}^0 = \mathbf{T}\mathbf{Q}^*$ and $\mathbf{X}^0 = \mathbf{D}\mathbf{Q}^*$, we further represent (\ref{eq:united_model}) as
\begin{equation}
\begin{aligned}
\begin{bmatrix}
\mathbf{P}^0 \\
\mathbf{X}^0
\end{bmatrix}
=
\begin{bmatrix}
\mathbf{T} \\
\mathbf{D}
\end{bmatrix}\mathbf{Q}^*.
\label{eq:px_model}
\end{aligned}
\end{equation}
It is clear that
\begin{equation}
\begin{aligned}
\mathrm{rank}(
\begin{bmatrix}
\mathbf{P}^0 ;
\mathbf{X}^0
\end{bmatrix})
& \leq \min \left(\mathrm{rank}(\begin{bmatrix}
\mathbf{T} ;
\mathbf{D}
\end{bmatrix}),
\mathrm{rank}(\mathbf{Q}^*)\right)
\\
& \leq \min \left(c + m , \sum_{k=1}^c n_k, N\right).
\end{aligned}
\end{equation}
Since the representatives in $\mathbf{D}$ only account for a subset of data samples, it is clear that $\sum_{k=1}^c n_k \leq N$. Therefore,
\begin{equation}
\begin{aligned}
\mathrm{rank}(
\begin{bmatrix}
\mathbf{P}^0 ;
\mathbf{X}^0
\end{bmatrix})
\leq \min \left(c + m , \sum_{k=1}^c n_k\right) \leq \sum_{k=1}^c n_k.
\end{aligned}
\end{equation}
In the case of $N \gg  \sum_{k=1}^c n_k$, the rank of $[\mathbf{P}^0; \mathbf{X}^0]$ is relatively smaller than $N$. From the above rank analysis and (\ref{eq:united_model}), we arrive at the following proposition:
\newtheorem{prop}{Proposition}
\begin{prop}
In the absence of noise, the heterogeneous feature matrix $\mathbf{H}_{obs}$ is low-rank.
\label{th:optimalPD}
\end{prop}
Note that a similar result is also reported in \cite{Cabral2014} without making the convex hull assumption.

\subsection{Matrix Completion for Ambiguity Resolution} \label{subsec:ambiguous_LR}
According to (\ref{eq:P_recover}), the true labeling matrix $\mathbf{P}^0$ can be recovered if $\mathbf{D}$ and $\mathbf{Q}^*$ are available. Nevertheless, obtaining $\mathbf{D}$ and $\mathbf{Q}^*$ based on the observed $\mathbf{P}$ and $\mathbf{X}$ is intractable by solving a matrix decomposition problem
\begin{equation}
\begin{aligned}
\min_{\mathbf{T}, \mathbf{D}, \mathbf{Q}} \left\|
\begin{bmatrix}
\mathbf{P} \\
\mathbf{X}
\end{bmatrix}
-
\begin{bmatrix}
\mathbf{T} \\
\mathbf{D}
\end{bmatrix}\mathbf{Q}
\right\|_F^2,
\end{aligned}\label{eqn:MatrixDecomp}
\end{equation}
subject to the conditions specified in (\ref{eq:convexhull_cat})-(\ref{eq:px_model}).
Following \cite{Goldberg2010}, we propose to resolve the ambiguity by recovering the underlying low-rank structure of the heterogeneous feature matrix.
Hence, we transform the matrix decomposition problem to a matrix completion problem.
For ease of presentation, we start with solving a label assignment problem assuming that $\mathbf{X}$ is noise-free, i.e. $\mathbf{X} = \mathbf{X}^0$. The predicted labeling matrix $\mathbf{Y}$ can be estimated by solving the following rank minimization problem
\begin{equation}
\begin{aligned}
 \min_{\mathbf{Y}, \mathbf{E}_P} \,\, & \mathrm{rank} \left(\begin{bmatrix}
\mathbf{Y} \\
\mathbf{X}^0
\end{bmatrix}\right) \\
 \text{s.t.} & \;
\begin{bmatrix}
\mathbf{Y} \\
\mathbf{X}^0
\end{bmatrix}= \begin{bmatrix}
\mathbf{P} \\
\mathbf{X}^0
\end{bmatrix} - \begin{bmatrix}
\mathbf{E}_P \\
\mathbf{0}
\end{bmatrix},\\
& \mathbf{y}_j \in \{\mathbf{v}_1, \mathbf{v}_2, \dots, \mathbf{v}_c\}, j = 1, 2, \dots, N,\\
& y_{i,j} = 0 \,\, \mathrm{if}  \,\, i \notin L_j \,\, \forall j. \\
\end{aligned}\label{eqn:Hard_NoiseFree}
\end{equation}
The problem is to complete the labeling matrix $\mathbf{Y}$ via pursuing a low-rank matrix $\begin{bmatrix}
\mathbf{Y} ;
\mathbf{X}^0
\end{bmatrix}$ subject to constraints given by the ambiguous labels. The first constraint defines the feasible region of label assignment and the second constraint implies that an instance can only be labeled among its candidate labels.
We cannot guarantee that the optimal solution to (\ref{eqn:Hard_NoiseFree}) always yields a perfect recovery of ambiguous labeling such that $\mathbf{Y}^*= \mathbf{P}^{0}$. Several factors contribute to our inability to resolve the ambiguity. For instance, if label $1$ is consistently present in the candidate labeling set of each instance, assigning $\mathbf{v}_1$ for each column vector of $\mathbf{Y}$ yields a trivial solution. This issue is also addressed in \cite{Cour2011}, as learning from instances associated with two consistently co-occurring labels is impossible.

Note that $\mathbf{Y}^*= \mathbf{P}^{0}$ is one of the possible optimal solutions to (\ref{eqn:Hard_NoiseFree}). The solution may not be unique if any one of the instances belongs to more than one convex hull, i.e. the convex hulls from different classes overlap with each other. Hence, an instance can be ideally decomposed from either one of the convex hulls without further changing the rank of $[\mathbf{Y} ; \mathbf{X}^0]$.
Nevertheless, it is our intention to seek $\mathbf{Y} = \mathbf{P}^0$ by solving (\ref{eqn:Hard_NoiseFree}) with the understanding that 1) the ambiguous labeling carries rational information, and 2) the feature is sufficiently discriminative such that data lies in the feature subspace where convex hulls of each class are separable \cite{Cover1965}.

\begin{figure}
\centering
\includegraphics[trim= 125 200 150 105,clip,width=0.5\textwidth]{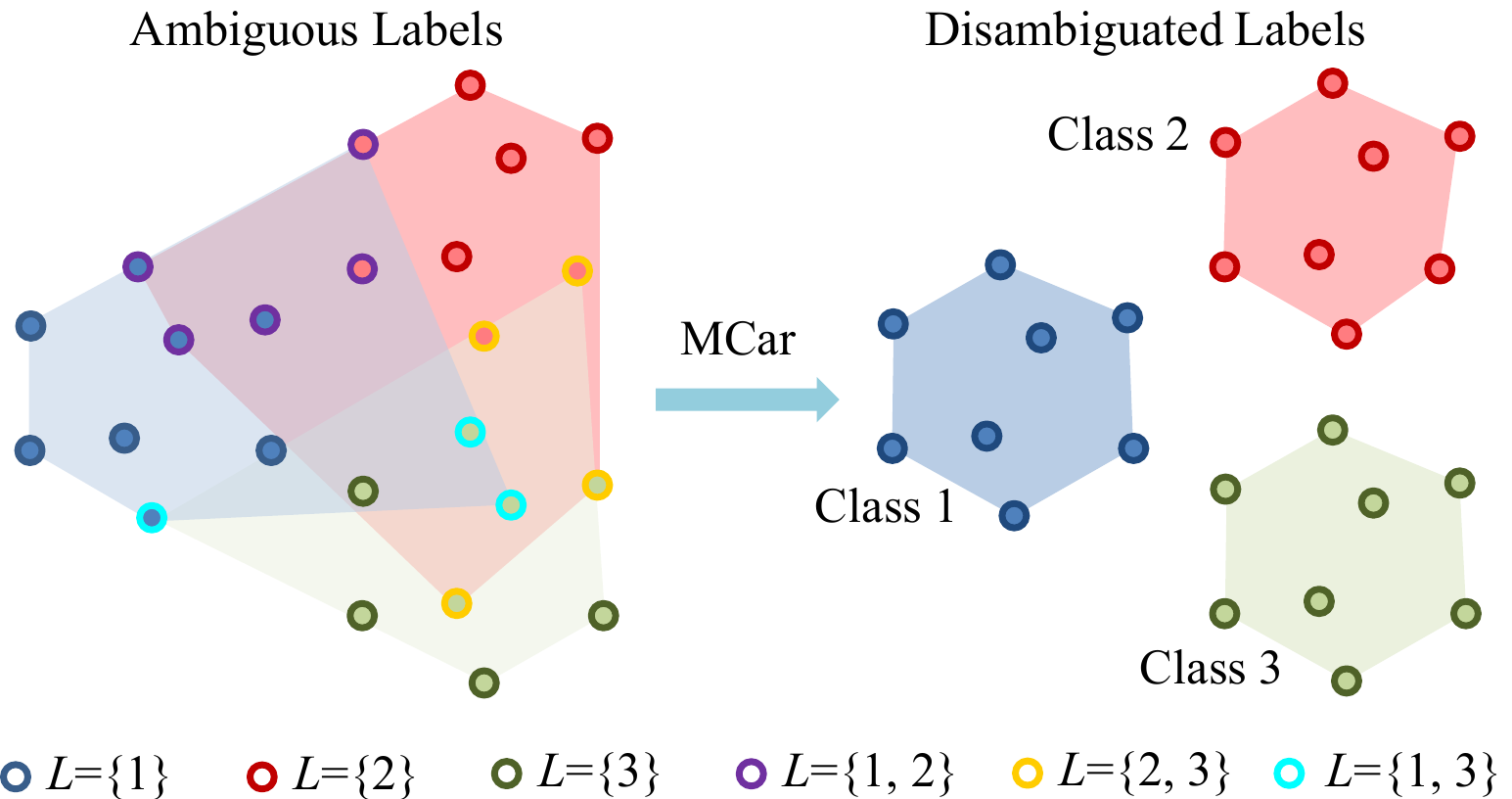}
\caption{MCar reassigns labels for those ambiguously labeled instances such that instances of the same subjects cohesively form potentially-separable convex hulls.
The vertices of each convex hull are the representatives of each class, forming $\mathbf{D}_k$.
The interior and outline of the circles are color-coded to represent three different classes and various ambiguous labels, respectively.}
\label{fig:convexhull}
\end{figure}

Figure \ref{fig:convexhull} illustrates the geometric interpretation of MCar using the convex hull representation. When each element in the candidate labeling set is trivially treated as the true label, the convex hulls of each class are erroneously expanded and the low-rank assumption of $\begin{bmatrix}
\mathbf{Y} ;
\mathbf{X}^0
\end{bmatrix}$ does not hold.
MCar exploits the underlying low-rank structure of $\begin{bmatrix}
\mathbf{Y} ;
\mathbf{X}^0
\end{bmatrix}$, which is equivalent to reassigning the labels for those ambiguously labeled instances such that instances of the same class cohesively form a convex hull. Hence, each over-expanded convex hull shrinks to its actual contour, and the convex hulls become potentially separable. This is essentially different from discriminative ambiguous learning methods that construct the hyperplane between ambiguously labeled instances by minimizing the ambiguous loss.

\begin{figure}[tb]\center
    \includegraphics[trim= 125 125 10 20,clip,width=0.5\textwidth]{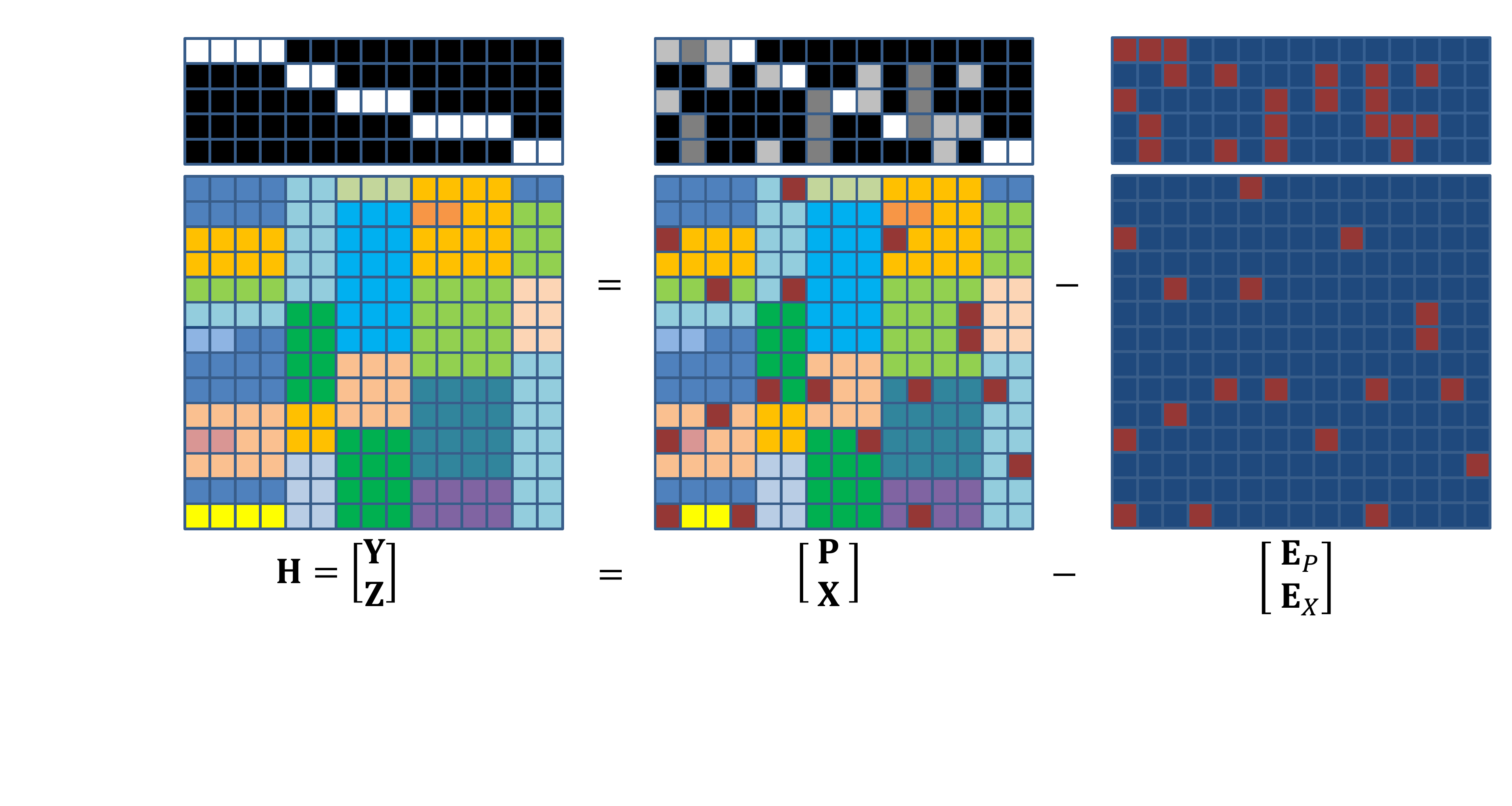}
    \caption{Ideal decomposition of the heterogeneous feature matrix using MCar. The underlying low-rank structure and the ambiguous labeling are recovered simultaneously.}
    \label{fig:LRM}
\end{figure}

When data is contaminated by sparse errors, the optimization problem in (\ref{eqn:Hard_NoiseFree}) can be reformulated as
\begin{equation}
\begin{aligned}
 \min_{\mathbf{H}, \mathbf{E}_X, \mathbf{E}_P} \,\, & \mathrm{rank} (\mathbf{H}) + \lambda \|\mathbf{E}_X\|_0  \\
 \text{s.t.} & \;
{\mathbf{H}}= \begin{bmatrix}
\mathbf{Y} \\
\mathbf{Z}
\end{bmatrix}= \begin{bmatrix}
\mathbf{P} \\
\mathbf{X}
\end{bmatrix} - \begin{bmatrix}
\mathbf{E}_P \\
\mathbf{E}_X
\end{bmatrix},\\
& \mathbf{y}_j \in \{\mathbf{v}_1, \mathbf{v}_2, \dots, \mathbf{v}_c\}, j = 1, 2, \dots, N,\\
& y_{i,j} = 0 \,\, \mathrm{if}  \,\, i \notin L_j \,\, \forall j, \\
\end{aligned}\label{eqn:Hard}
\end{equation}
where $\mathbf{H}$ is the heterogeneous feature matrix in the absence of noise, and $\mathbf{Z}$ is the recovered feature matrix. The parameter $\lambda \in \mathbb{R}_+$ controls the rank of $\mathbf{H}$ and the sparsity of noise. The objective is to assign the predicted label $\mathbf{Y}$ and extract the sparse noise of $\mathbf{X}$ in pursuit of a low-rank $\mathbf{H}$. Figure \ref{fig:LRM} illustrates the ideal decomposition of the heterogeneous feature matrix, where the underlying low-rank structure and the ambiguous labels are recovered simultaneously.

As (\ref{eqn:Hard}) is a combinatorial optimization problem, we relax each column vector of $\mathbf{Y}$ in probability simplex in $\mathbb{R}^c$.
The original formulation can be rewritten as
\begin{equation}
\begin{aligned}
 \min_{\mathbf{H}, \mathbf{E}_X, \mathbf{E}_P} \,\, & \mathrm{rank} (\mathbf{H}) + \lambda \|\mathbf{E}_X\|_0  +  \gamma \|\mathbf{Y}\|_0\\
 \text{s.t.} & \;
{\mathbf{H}}= \begin{bmatrix}
\mathbf{Y} \\
\mathbf{Z}
\end{bmatrix}= \begin{bmatrix}
\mathbf{P} \\
\mathbf{X}
\end{bmatrix} - \begin{bmatrix}
\mathbf{E}_P \\
\mathbf{E}_X
\end{bmatrix},\\
& \mathbf{1}^T_c \mathbf{Y} = \mathbf{1}^T_N, \,\, \mathbf{Y} \in \mathbb{R}_+^{c \times N}, \\
& y_{i,j} = 0 \,\, \mathrm{if}  \,\, i \notin L_j \,\, \forall j, \\
\end{aligned}\label{eqn:Y_relaxed1}
\end{equation}
where $\gamma \in \mathbb{R}_+$ encourages the sparsity of $\mathbf{Y}$ such that the original discrete feasible region can be well approximated.
From the perspective of convex hull representation, such relaxation allows each instance to be represented from more than one set of representative matrix $\mathbf{D}_k$, while it will be penalized by the non-sparsity of $\mathbf{Y}$.
Consequently, the predicted label of instance $j$ can be obtained as
\begin{equation}
\begin{aligned}
\hat{l}_j = \arg \max_{i \in L_j} \,\,  y_{i,j}.
\end{aligned}\label{eqn:ypredict}
\end{equation}

\subsection{Ambiguously Labeled Data with Labeling Imbalance}

Class imbalance often leads to performance degradation in support vector machine (SVM) as a majority class with abundant training samples can bias the decision boundary toward a minority class with fewer training samples.
Analogously, MCar may suffer from labeling imbalance when a majority label is frequently present among the candidate labels in the ambiguously labeled data.
When we resolve the ambiguity using (\ref{eqn:Y_relaxed1}), the heterogeneous feature vectors associated with a majority label are more likely to dominate the low-rank approximation of the heterogeneous matrix than those associated with a minor label. Hence, the recovered soft labeling matrix will bias toward soft labeling vectors associated with majority labels.

Class-weighted SVM applies unequal weighting to the cost function of different classes to mitigate the class imbalance \cite{Veropouloscts1999}. Hence, instances from the minority label will be better emphasized than those from the dominant label to establish an objective decision boundary.
However, the concept of class-weighted SVM cannot be directly applied to MCar to deal with label imbalances since each instance is not labeled as a particular class in the ambiguously labeled data. Without the knowledge of the true labels, we formulate the instance-weighted objective function of (\ref{eqn:MatrixDecomp}) as
\begin{equation}
\begin{aligned}
\min_{\mathbf{T}, \mathbf{D}, \mathbf{Q}} \sum_{j=1}^N \eta_j \left\|
\begin{bmatrix}
\mathbf{p}_j \\
\mathbf{x}_j
\end{bmatrix}
-
\begin{bmatrix}
\mathbf{T} \\
\mathbf{D}
\end{bmatrix}\mathbf{q}_j
\right\|_F^2
,
\end{aligned}\label{eqn:W_MatrixDecomp}
\end{equation}
where $\eta_j$ is the instance weight of the $j^{th}$ instance. In order to balance the square errors contributed by each class in (\ref{eqn:W_MatrixDecomp}), we aim to set instance weight $\eta_j$ as $1/N_{l_j}$, where $N_{l_j}$ is the number of the instances from the $l_j$ class. Nevertheless, assigning a class weight for each instance is not feasible in the ambiguously labeled data since the true label $l_j$ is not explicitly known. Moreover, $N_{i}$ is intractable since the data is not explicitly labeled. Hence, we propose to set the instance weight as
\begin{equation}
\begin{aligned}
\eta_j = \frac{1}{\sum_{i=1}^c p_{i,j} \hat{N}_i},
\end{aligned}
\end{equation}
where
\begin{equation}
\begin{aligned}
\hat{N}_i =  \sum_{j=1}^N p_{i,j} \label{eqn:est_sub_dist}
\end{aligned}
\end{equation}
is the estimated number of instances of the $i^{th}$ class. The estimated number of instances of the $i^{th}$ class accumulates the soft labeling scores corresponding to the $i^{th}$ class across all instances. With soft labeling vector $\mathbf{p}_{j}$, we can compute the effective number of instances of the class that the $j^{th}$ instance belongs to by $\sum_{i=1}^c p_{i,j} \hat{N}_i$.
Hence, our proposed weighting scheme is eligible to compute the effective class weight of each ambiguously labeled instance even though the knowledge of true label is not available.
The design of the instance weight is not unique, and readers may refer to \cite{Lin2002svm,He2009lfi} for modeling the instance weight with respect to various objectives.

For ease of presentation, we reformulate (\ref{eqn:W_MatrixDecomp}) as

\begin{equation}
\begin{aligned}
\min_{\mathbf{T}, \mathbf{D}, \mathbf{Q}} \left\|
\begin{bmatrix}
\mathbf{P} \\
\mathbf{X}
\end{bmatrix} \mathbf{W}
-
\begin{bmatrix}
\mathbf{T} \\
\mathbf{D}
\end{bmatrix} \mathbf{Q} \mathbf{W}
\right\|_F^2
,
\end{aligned}
\end{equation}
where
\begin{equation}
\begin{aligned}
\mathbf{W} = \sqrt{\diag(\mathbf{1}_N^T \mathbf{P}^T \mathbf{P})}^{-1}
\end{aligned}\label{eqn:W_matrix}
\end{equation}
is a diagonal weighting matrix with $w_{j,j} = \sqrt{\eta_j}$.
As post-multiplying $\mathbf{W}$ does not increase the rank of a matrix, we claim that \emph{Proposition \ref{th:optimalPD}} also applies to the weighted heterogeneous feature matrix $\mathbf{H}_{obs} \mathbf{W} = [\mathbf{P};\mathbf{X}] \mathbf{W}$. We propose the weighted MCar (WMCar) by generalizing (\ref{eqn:Y_relaxed1}) as
\begin{equation}
\begin{aligned}
 \min_{\mathbf{H}, \mathbf{E}_X, \mathbf{E}_P} \,\, & \mathrm{rank} (\mathbf{H}\mathbf{W}) + \lambda \|\mathbf{E}_X \mathbf{W}\|_0  +  \gamma \|\mathbf{Y} \mathbf{W}\|_0\\
 \text{s.t.} & \;
{\mathbf{H}\mathbf{W}}= \begin{bmatrix}
\mathbf{Y} \mathbf{W}\\
\mathbf{Z} \mathbf{W}
\end{bmatrix}= \begin{bmatrix}
\mathbf{P} \mathbf{W} \\
\mathbf{X} \mathbf{W}
\end{bmatrix} - \begin{bmatrix}
\mathbf{E}_P \mathbf{W}\\
\mathbf{E}_X \mathbf{W}
\end{bmatrix},\\
& \mathbf{1}^T_c \mathbf{Y}\mathbf{W} = \mathbf{1}^T_N\mathbf{W}, \,\, \mathbf{Y}\mathbf{W} \in \mathbb{R}_+^{c \times N}, \\
& y_{i,j} = 0 \,\, \mathrm{if}  \,\, i \notin L_j \,\, \forall j. \\
\end{aligned}\label{eqn:Y_relaxed1_W1}
\end{equation}
Let $\bar{\mathbf{H}}_{obs} = \mathbf{H}_{obs} \mathbf{W}$, $\bar{\mathbf{H}} = \mathbf{H} \mathbf{W}$, and $\bar{\mathbf{E}} = \mathbf{E} \mathbf{W}$, we reformulate (\ref{eqn:Y_relaxed1_W1}) as
\begin{equation}
\begin{aligned}
 \min_{\bar{\mathbf{H}}, \bar{\mathbf{E}}_P, \bar{\mathbf{E}}_X} \,\, & \mathrm{rank} (\bar{\mathbf{H}}) + \lambda \|\bar{\mathbf{E}}_X \|_0  +  \gamma \|\bar{\mathbf{Y}} \|_0\\
 \text{s.t.} & \;
{\bar{\mathbf{H}}}= \begin{bmatrix}
\bar{\mathbf{Y}}\\
\bar{\mathbf{Z}}
\end{bmatrix}= \begin{bmatrix}
\bar{\mathbf{P}}\\
\bar{\mathbf{X}}
\end{bmatrix} - \begin{bmatrix}
\bar{\mathbf{E}}_P \\
\bar{\mathbf{E}}_X
\end{bmatrix},\\
& \mathbf{1}^T_c \bar{\mathbf{Y}} = \mathbf{1}^T_N\mathbf{W}, \,\, \bar{\mathbf{Y}} \in \mathbb{R}_+^{c \times N}, \\
& \bar{y}_{i,j} = 0 \,\, \mathrm{if} \,\, i \notin L_j \,\, \forall j. \\
\end{aligned}\label{eqn:Y_relaxed1_W2}
\end{equation}
The predicted label can be retrieved from $\mathbf{Y} = \bar{\mathbf{Y}} \mathbf{W}^{-1}$ using (\ref{eqn:ypredict}).
Interestingly, the instance-weighted MCar is equivalent to executing MCar with the weighted heterogeneous feature matrix.
A larger weight on the heterogeneous feature vectors associated with minority labels provides those instances a stronger impact in the low-rank approximation of the heterogeneous matrix, and thus the labeling imbalance can be compensated.
As (\ref{eqn:Y_relaxed1}) is generalized by (\ref{eqn:Y_relaxed1_W2}) in consideration of labeling imbalance, WMCar is identical to MCar in the special case of $\mathbf{W} = \mathbf{I}$.

\begin{algorithm}
  \caption{The optimization algorithm for WMCar (\ref{eqn:Y_relaxed_augR})}
  \begin{algorithmic}[1]
    \Require
     $\mathbf{P} \in \mathbb{R}^{c \times N}$, $\mathbf{X} \in \mathbb{R}^{m \times N}$, $\mathbf{W} \in \mathbb{R}^{N \times N}$, $L_j \, \forall j$, $\lambda$, and $\gamma$.\\
    \textbf{Initialization:}\\ $\bar{\mathbf{P}} = \mathbf{P} \mathbf{W}$, $\bar{\mathbf{X}} = \mathbf{X} \mathbf{W}$, $\bar{\mathbf{H}}_{obs} = [\bar{\mathbf{P}}; \bar{\mathbf{X}}] $; \\
    $\bar{\mathbf{Y}}=\mathbf{0}$, $\bar{\mathbf{Z}} = \mathbf{0}$, $\mu > 0, \mu_{\max} > 0$,  $\rho > 1$, $\mathbf{\Lambda} = [\mathbf{\Lambda}_P; \mathbf{\Lambda}_X] = \bar{\mathbf{H}}_{obs} / \| \bar{\mathbf{H}}_{obs}\|_2$;
    \While {not converged}
      \State $\bar{\mathbf{E}}_P = \bar{\mathbf{P}} - \mathcal{S}_{\gamma \mu^{-1}}[ \bar{\mathbf{Y}} - \mu^{-1} \mathbf{\Lambda}_P]$;
      \State $\bar{\mathbf{E}}_X = \mathcal{S}_{\lambda \mu^{-1}}[ \bar{\mathbf{X}} - \bar{\mathbf{Z}} + \mu^{-1} \mathbf{\Lambda}_X]$;
      \State $(\mathbf{U}, \mathbf{\Sigma}, \mathbf{V}) = \mathrm{svd}\left(\bar{\mathbf{H}}_{obs} - \bar{\mathbf{E}} + \mu^{-1} \mathbf{\Lambda}\right)$;
      \State $\bar{\mathbf{H}} = \mathbf{U} \mathcal{S}_{\mu^{-1}}[\mathbf{\Sigma}] \mathbf{V}^T $;
      \State $\mathbf{\Lambda} = \mathbf{\Lambda} + \mu \left(\bar{\mathbf{H}}_{obs} - \bar{\mathbf{H}} - \bar{\mathbf{E}} \right)$;
      \State $\mu = \min(\rho \mu , \mu_{\max} )$;
      \State \textbf{Project} $\bar{\mathbf{Y}}$:
      \LineComment{Line: 13: Projection for (\ref{cond:al1})}
      \State $\bar{y}_{i,j} = 0 \,\, \mathrm{if}   \,\, i \notin L_j \,\, \forall j$;
      \LineComment{Line: 15-16: Projection for (\ref{cond:probsim1})}
      \State $\bar{\mathbf{Y}} = \max(\bar{\mathbf{Y}}, 0)$;
      \State $\bar{\mathbf{y}}_j = w_{j,j} \,\bar{\mathbf{y}}_j / \|\bar{\mathbf{y}}_j\|_1,$ $\forall j$;
      \EndWhile\\
    $\mathbf{H} = \bar{\mathbf{H}} \mathbf{W}^{-1}$, $\mathbf{E} = \bar{\mathbf{E}} \mathbf{W}^{-1}$
    \Ensure
        $(\mathbf{H}, \mathbf{E})$
    \end{algorithmic}\label{alg:alm_W}
\end{algorithm}

\section{Optimization}\label{MCar_sec:opt}
The augmented Lagrangian method (ALM) has been extensively used for solving low-rank problems \cite{Candes2011,Lin2009}.
In this section, we propose to incorporate the ALM with the projection step discussed in \cite{Goldberg2010,Cabral2011} to solve the optimization problem of WMCar.

In order to decouple $\bar{\mathbf{Y}}$ in the first and third terms of the objective function in (\ref{eqn:Y_relaxed1_W2}), we replace
$\|\bar{\mathbf{Y}}\|_0$ with $\|\bar{\mathbf{P}} -\bar{\mathbf{E}}_P\|_0$ and rewrite (\ref{eqn:Y_relaxed1_W2}) as
\begin{equation}
\begin{aligned}
 \min_{\bar{\mathbf{H}}, \bar{\mathbf{E}}_X, \bar{\mathbf{E}}_P} \,\, & \mathrm{rank} (\bar{\mathbf{H}}) + \lambda \|\bar{\mathbf{E}}_X\|_0  +  \gamma \|\bar{\mathbf{P}} -\bar{\mathbf{E}}_P\|_0\\
 \text{s.t.} & \;
{\bar{\mathbf{H}}}= \begin{bmatrix}
\bar{\mathbf{Y}} \\
\bar{\mathbf{Z}}
\end{bmatrix}= \begin{bmatrix}
\bar{\mathbf{P}} \\
\bar{\mathbf{X}}
\end{bmatrix} - \begin{bmatrix}
\bar{\mathbf{E}}_P \\
\bar{\mathbf{E}}_X
\end{bmatrix},\\
& \mathbf{1}^T_c \bar{\mathbf{Y}}  = \mathbf{1}^T_N \mathbf{W}, \,\, \bar{\mathbf{Y}} \in \mathbb{R}_+^{c \times N}, \\
& \bar{y}_{i,j} = 0 \,\, \mathrm{if} \,\, i \notin L_j \,\, \forall j. \\
\end{aligned}\label{eqn:Y_relaxed2}
\end{equation}
Following the procedure of ALM, we relax the first constraint in (\ref{eqn:Y_relaxed2}) and reformulate it as
\begin{equation}
\begin{aligned}
 \min_{\bar{\mathbf{H}},\bar{\mathbf{E}}, \mathbf{\Lambda}, \mu} \,\,& \ell  (\bar{\mathbf{H}},\bar{\mathbf{E}}, \mathbf{\Lambda}, \mu) \\
 \text{s.t.}
&\,\, \; \mathbf{1}^T_c \bar{\mathbf{Y}} = \mathbf{1}^T_N \mathbf{W}, \,\, \bar{\mathbf{Y}} \in \mathbb{R}_+^{c \times N}, \\
& \bar{y}_{i,j} = 0 \,\, \mathrm{if} \,\, i \notin L_j \,\, \forall j, \\
\end{aligned}\label{eqn:Y_relaxed_aug}
\end{equation}
where  $\mu \in \mathbb{R}_+$ and $\mathbf{\Lambda} \in \mathbb{R}^{(c+m) \times N}$.
The Lagrangian is expressed as
\begin{equation}
\begin{aligned}
\ell  (\bar{\mathbf{H}},\bar{\mathbf{E}}, \mathbf{\Lambda},& \mu) = \mathrm{rank} (
\bar{\mathbf{H}}) + \lambda \|\bar{\mathbf{E}}_X\|_0  +  \gamma \|\bar{\mathbf{P}} -\bar{\mathbf{E}}_P\|_0 \\
& + \left\langle  \mathbf{\Lambda} ,
\bar{\mathbf{H}}_{obs} - \bar{\mathbf{H}}
-
\bar{\mathbf{E}}
\right\rangle  + \frac{\mu}{2} \left\|
\bar{\mathbf{H}}_{obs} - \bar{\mathbf{H}}  -
\bar{\mathbf{E}}
 \right\|_F^2.
\end{aligned}\label{eqn:L}
\end{equation}
In order to make the optimization problem feasible, we approximate the rank with the nuclear norm and the $\ell_{0}$ norm with the $\ell_{1}$ norm \cite{Candes_matrixcompletion}.  Thus, we solve the following formulation as the convex surrogate of (\ref{eqn:Y_relaxed_aug})
\begin{align}
\min_{\bar{\mathbf{H}},\bar{\mathbf{E}}, \mathbf{\Lambda}, \mu} \,\,& \ell_R  (\bar{\mathbf{H}},\bar{\mathbf{E}}, \mathbf{\Lambda}, \mu) \label{eqn:Y_relaxed_augR} \\
\text{s.t.}
&\,\, \; \mathbf{1}^T_c \bar{\mathbf{Y}} = \mathbf{1}^T_N \mathbf{W}, \,\, \bar{\mathbf{Y}} \in \mathbb{R}_+^{c \times N}, \label{cond:probsim1} \\
& \bar{y}_{i,j} = 0 \,\, \mathrm{if} \,\, i \notin L_j \,\, \forall j,\label{cond:al1}
\end{align}
where the Lagrangian is represented as
\begin{equation}
\begin{aligned}
\ell_R   (\bar{\mathbf{H}},\bar{\mathbf{E}}, \mathbf{\Lambda},& \mu) =
 \left\| \bar{\mathbf{H}} \right\|_* + \lambda \|\bar{\mathbf{E}}_X\|_1  +  \gamma \|\bar{\mathbf{P}} - \bar{\mathbf{E}}_P\|_1 \\
 +& \left \langle  \mathbf{\Lambda} ,
\bar{\mathbf{H}}_{obs} -
\bar{\mathbf{H}}
-
\bar{\mathbf{E}}
 \right \rangle  + \frac{\mu}{2} \left\|
\bar{\mathbf{H}}_{obs} - \bar{\mathbf{H}}
-
\bar{\mathbf{E}}
\right\|_F^2.
\end{aligned}\label{eqn:LR}
\end{equation}
The ALM operates in the sense that $\bar{\mathbf{H}}$, $\bar{\mathbf{E}}_P$, and $\bar{\mathbf{E}}_X$ can be solved alternately by fixing other variables. In each iteration, we employ a similar projection technique used in \cite{Goldberg2010,Cabral2011} to enforce $\bar{\mathbf{Y}}$ to be feasible.
The entire procedure for solving (\ref{eqn:Y_relaxed_augR}) is summarized in Algorithm \ref{alg:alm_W}, and the details of the optimization algorithm are presented in the following paragraphs.

\subsection{Solving for $\bar{\mathbf{E}}_P$}
To update $\bar{\mathbf{E}}_P$, we fix $\bar{\mathbf{H}}$, $\bar{\mathbf{E}}_X$, $\mathbf{\Lambda}$ and $\mu$ obtained in the previous iteration.
Hence, the problem for updating $\bar{\mathbf{E}}_P$ can be solved by first computing
{
\begin{equation}
\begin{aligned}
\bar{\mathbf{E}}_P^* = \arg\!\min_{\bar{\mathbf{E}}_P} & \,\,  \gamma  \|\bar{\mathbf{P}} - \bar{\mathbf{E}}_P\|_1
 + \left \langle  \mathbf{\Lambda}_P , \bar{\mathbf{P}} - \bar{\mathbf{Y}} - \bar{\mathbf{E}}_P
 \right \rangle \\
&+ \frac{\mu}{2}  \left\|
\bar{\mathbf{P}} - \bar{\mathbf{Y}} - \bar{\mathbf{E}}_P
 \right\|_F^2.\label{eqn:ep}
\end{aligned}
\end{equation}}
For the ease of derivation, we let $\bar{\mathbf{B}} =  \bar{\mathbf{P}} - \bar{\mathbf{E}}_P$ and update $\bar{\mathbf{B}}$ as surrogate. We can reformulate (\ref{eqn:ep}) as
{
\begin{equation}
\begin{aligned}
\bar{\mathbf{B}}^* &= \arg\!\min_{\bar{\mathbf{B}}} \,\,  \gamma  \|\bar{\mathbf{B}}\|_1
 + \left \langle  \mathbf{\Lambda}_P , \bar{\mathbf{B}} - \bar{\mathbf{Y}}
 \right \rangle + \frac{\mu}{2}  \left\|
\bar{\mathbf{B}} - \bar{\mathbf{Y}}
 \right\|_F^2, \\
&= \arg\!\min_{\bar{\mathbf{B}}} \,\,  \gamma  \|\bar{\mathbf{B}}\|_1  + \frac{\mu}{2} \| \bar{\mathbf{B}} - \bar{\mathbf{Y}}   + \mu^{-1}\mathbf{\Lambda}_P  \|_F^2,\\
&= \arg\!\min_{\bar{\mathbf{B}}} \,\,  \gamma  \|\bar{\mathbf{B}}\|_1  + \frac{\mu}{2} \|   \bar{\mathbf{Y}}  - \mu^{-1}\mathbf{\Lambda}_P - \bar{\mathbf{B}} \|_F^2. \label{eqn:ep_last}
\end{aligned}
\end{equation}}
Using the subgradient of (\ref{eqn:ep_last}), we can obtain the closed-form solution for updating $\mathbf{B}$
\begin{align}
\bar{\mathbf{B}}^* = \mathcal{S}_{ \gamma  \mu^{-1}}[ \bar{\mathbf{Y}} - \mu^{-1} \mathbf{\Lambda}_P].
\end{align}
Consequently, we can update $\bar{\mathbf{E}}_P$ as
\begin{align}
\bar{\mathbf{E}}_P^* = \bar{\mathbf{P}} - \bar{\mathbf{B}}^* = \bar{\mathbf{P}} - \mathcal{S}_{ \gamma  \mu^{-1}}[ \bar{\mathbf{Y}} - \mu^{-1} \mathbf{\Lambda}_P].
\end{align}

\subsection{Solve $\bar{\mathbf{E}}_X$}
To update $\bar{\mathbf{E}}_X$, we fix $\bar{\mathbf{H}}$, $\bar{\mathbf{E}}_P$, $\mathbf{\Lambda}$ and $\mu$ obtained in the previous iteration.
Thus, the problem for updating $\bar{\mathbf{E}}_X$ can be solved by
\begin{align}
\nonumber \bar{\mathbf{E}}_X^* &= \arg\!\min_{\bar{\mathbf{E}}_X} \,\,  \lambda \|\bar{\mathbf{E}}_X\|_1
 + \left \langle  \mathbf{\Lambda}_X , \bar{\mathbf{X}} - \bar{\mathbf{Z}} - \bar{\mathbf{E}}_X
 \right \rangle \\
 \nonumber & + \frac{\mu}{2}  \left\|
\bar{\mathbf{X}} - \bar{\mathbf{Z}} - \bar{\mathbf{E}}_X
 \right\|_F^2, \\
&= \arg\!\min_{\bar{\mathbf{E}}_X} \,\, \lambda \|\bar{\mathbf{E}}_X\|_1  + \frac{\mu}{2} \| \bar{\mathbf{X}} - \bar{\mathbf{Z}}  + \mu^{-1}\mathbf{\Lambda}_X - \bar{\mathbf{E}}_X \|_F^2. \label{eqn:ex_last}
\end{align}
Using the subgradient of (\ref{eqn:ex_last}), we can obtain the closed-form solution for updating $\mathbf{E}_X$
\begin{align}
\bar{\mathbf{E}}_X^* = \mathcal{S}_{\lambda \mu^{-1}}[ \bar{\mathbf{X}} - \bar{\mathbf{Z}} + \mu^{-1} \mathbf{\Lambda}_X].
\end{align}

\subsection{Solve $\bar{\mathbf{H}}$}
To update $\bar{\mathbf{H}}$, we fix $\bar{\mathbf{E}}_P$, $\bar{\mathbf{E}}_X$, $\mathbf{\Lambda}$ and $\mu$ obtained in the previous iteration. The feasible region of $\bar{\mathbf{Y}}$ in $\bar{\mathbf{H}}$ is currently not considered but will be handled in the projection step of $\bar{\mathbf{Y}}$ (Section \ref{subsec:y_step}).
Therefore, the problem for updating $\bar{\mathbf{H}}$ can be solved by
\begin{align}
\bar{\mathbf{H}}^* &= \arg\!\min_{\bar{\mathbf{H}}} \,\,  \| \bar{\mathbf{H}} \|_*
 + \left \langle  \mathbf{\Lambda} ,
\bar{\mathbf{H}}_{obs} -\bar{\mathbf{H}}
-
\bar{\mathbf{E}}
 \right \rangle \\
&+ \frac{\mu}{2}  \left\|
\bar{\mathbf{H}}_{obs} - \bar{\mathbf{H}}
 -
\bar{\mathbf{E}}
 \right\|_F^2, \\
&= \arg\!\min_{\bar{\mathbf{H}}} \,\, \| \bar{\mathbf{H}} \|_*  + \frac{\mu}{2} \| \mathbf{A}_H - \bar{\mathbf{H}} \|_F^2,
\end{align}
where $\mathbf{A}_H = \bar{\mathbf{H}}_{obs}  - \bar{\mathbf{E}} + \mu^{-1} \mathbf{\Lambda}$.
According to \cite{Cai2010}, the above problem can be solved by
\begin{align}
\bar{\mathbf{H}}^* = \mathbf{U} \mathcal{S}_{\mu^{-1}}[\mathbf{\Sigma}] \mathbf{V}^T,
\end{align}
where $\mathbf{\Sigma}$ can be obtained from the singular value decomposition (SVD) of $\mathbf{A}_H$ denoted as $(\mathbf{U}, \mathbf{\Sigma}, \mathbf{V}) = \mathrm{svd}\left(\mathbf{A}_H\right).$
Following the procedure in ALM, we can update $\mathbf{\Lambda}$ and  $\mu$ as
\begin{align}
       \mathbf{\Lambda} = \mathbf{\Lambda} + \mu \left(\bar{\mathbf{H}}_{obs} - \bar{\mathbf{H}} - \bar{\mathbf{E}} \right),
\end{align}
where $\mu = \min(\rho \mu , \mu_{\max} )$, in each iteration based on the updated $\bar{\mathbf{E}}_P$, $\bar{\mathbf{E}}_X$, and $\bar{\mathbf{H}}$.

\subsection{Project $\bar{\mathbf{Y}}$}  \label{subsec:y_step}
Since the SVD operation for solving $\bar{\mathbf{H}}$ does not always return a feasible $\bar{\mathbf{Y}}$, we use a projection technique similar to the one in \cite{Goldberg2010,Cabral2011} to enforce $\bar{\mathbf{Y}}$ to be feasible in each iteration. The projection involves two steps. First, we enforce those entries of $\bar{\mathbf{Y}}$ that do not correspond to the candidate labels to be zeros since the actual label only comes from the candidate labeling set provided by the ambiguous labels. Second, each column vector of $\mathbf{Y} = \bar{\mathbf{Y}} \mathbf{W}^{-1}$ is constrained to be in the probability simplex. As a result, we replace those negative entries in $\bar{\mathbf{Y}}$ with zeros and then normalize each column $\bar{\mathbf{y}}_j$ so that the summation of the entries in $\bar{\mathbf{y}}_j$ is equal to $w_{j,j}$.

\begin{algorithm}[htp!]
  \caption{The algorithm for WMCar-ICE}
  \begin{algorithmic}[1]
    \Require
     $\mathbf{P} \in \mathbb{R}^{c \times N}$, $\mathbf{X} \in \mathbb{R}^{m \times N}$, $L_j \,\, \forall j$.
    \While {$\mathcal{A} \neq \varnothing$ and within the maximum number of iterations}
      \State $\mathbf{W} = \sqrt{\diag(\mathbf{1}_N^T \mathbf{P}^T \mathbf{P})}^{-1} $;
      \State Obtain $\mathbf{Y}$ using WMCar (Algorithm \ref{alg:alm_W});
      \State Eliminate the least likely candidate in $L_j$, $j \in \mathcal{E}$ using (\ref{eqn:least_candidate})-(\ref{eqn:l_update});
      \LineComment{Line: \ref{line:iwmcar_p1}-\ref{line:iwmcar_p2}: Project $\mathbf{Y}$ to comply with $L_j, \forall j$}
      \State $y_{i,j} = 0$, if $ i \notin L_j$ $\forall j$;\label{line:iwmcar_p1}
      \State $\mathbf{y}_j = \mathbf{y}_j / \|\mathbf{y}_j\|_1$ $\forall j$;\label{line:iwmcar_p2}
      \State $\mathbf{P} \leftarrow \mathbf{Y}$;
    \EndWhile
    \Ensure
        $(\mathbf{H}, \mathbf{E})$
  \end{algorithmic}\label{alg:iwmcar}
\end{algorithm}

\section{Iterative Candidate Elimination for Ambiguity Resolution} \label{sec:ice}
According to (\ref{eqn:W_matrix}), the weighting matrix $\mathbf{W}$ of WMCar is a function of $\mathbf{P}$.
As WMCar resolves the label ambiguity in $\mathbf{P}$, the recovered soft labeling matrix $\mathbf{Y}$ can provide a better estimate of $\mathbf{W}$ than the original $\mathbf{P}$.
This motivates us to iteratively resolve the ambiguity by alternating between recovering $\mathbf{Y}$ and updating $\mathbf{W}$. Nevertheless, the performance of iterative WMCar is not steady as shown in Figure \ref{fig:WMCar_ICE_thre}. We propose WMCar with ICE (WMCar-ICE) to resolve the ambiguity by WMCar and then remove the least likely candidate labels in each iteration. The least likely candidate label of the $j^{th}$ instance is denoted as
\begin{align}
m(j)= \arg\!\min_{i \in L_j} \,\, y_{i,j}, \label{eqn:least_candidate}
\end{align}
and its corresponding soft labeling score is denoted as $y_{m(j),j}$.
As removing a candidate label, which is actually a true label, in the candidate set generates an irreversible error, we propose to iteratively remove a portion of the least likely candidate labels that have relatively low soft labeling scores than others.

Let $\mathcal{A}$ denote the set consisting of the indices of those instances that have more than one candidate label, which is represented as
\begin{align}
     \mathcal{A} =\{j  \,\, | \,\, |L_j| > 1, \forall j\}.
\end{align}
We define the elimination factor as $f_e$ ($ 0 \leq f_e \leq 1$), which accounts for the proportion of instances in $\mathcal{A}$ participating in the candidate elimination.
We construct a subset $\mathcal{E}$ of $\mathcal{A}$, which consists of entries that correspond to the smallest $f_e$ portion of $\{y_{m(j),j} | j \in \mathcal{A} \}$. We represent it as
\begin{align}
     \mathcal{E} =\{j  \,\, | \,\,  y_{m(j),j} \leq t,  j \in \mathcal{A} \}. \label{eqn:e_set}
\end{align}
Note that $t$ is automatically determined such that $|\mathcal{E}| = \ceil*{f_e \,\, |\mathcal{A}|}$.
Hence, we can update the candidate labeling sets by
\begin{align}
     L_j  \leftarrow L_j -  \{m(j)\}, \,\, j \in \mathcal{E}. \label{eqn:l_update}
\end{align}

We enforce the soft labeling matrix $\mathbf{Y}$ to comply with updated candidate labeling sets. We set $y_{i,j} = 0$, if $ i \notin L_j$ $\forall j$ and project each column vector of $\mathbf{Y}$ in the probability simplex. The original $\mathbf{P}$ will be replaced by $\mathbf{Y}$, which will serve as the input of WMCar in the next iteration.
The procedure of WMCar-ICE is summarized in Algorithm \ref{alg:iwmcar}.
Note that updating the weighting matrix $\mathbf{W}$ is an important step in WMCar-ICE since it adaptively adjusts the importance among instances based on the updated $\mathbf{Y}$ in the previous iteration.
This ICE procedure can be utilized by other ambiguous learning techniques that adopt the soft labeling input/output similar to that of WMCar.

\section{Labeling Constraints between Instances} \label{sec:lpc}
In practical applications, several ambiguously labeled instances can appear in the same venue.  As a result, pairwise relations between instances can be utilized to assist ambiguity resolution. For example, two persons in a news photo should not be identified as the same subject even though both of them are ambiguously labeled in the caption. Such prior knowledge can be easily incorporated by restricting the feasible region of the labeling matrix. Moreover, it is essential to handle the open set problem, where there are some instances whose identities never appear in the labels. These unrecognized instances can be treated as the null class.

In this section, we show how MCar's formulation can be extended to associate the identities in news photos when the names are provided in captions. We assume all the instances (face images) are collected from the $K$ groups (photos), and $G_k$ is the set of indices of the instances (face images) appearing in the $k^{th}$ group (photo). Note that instances (face images) from the same group (photo) share the same ambiguous labels provided by their associated caption. Without loss of generality, we assume that the $c^{th}$ class corresponds to the null class. Considering the prior knowledge, the original formulation given in (\ref{eqn:Y_relaxed1}) can be reformulated as
\begin{align}
 \min_{\mathbf{H}, \mathbf{E}_X, \mathbf{E}_P} \,\, & \mathrm{rank} (\mathbf{H}) + \lambda \|\mathbf{E}_X\|_0  +  \gamma \|\mathbf{Y}\|_0  \label{eqn:HardExt} \\
 \text{s.t.} & \;
{\mathbf{H}}= \begin{bmatrix}
\mathbf{Y} \\
\mathbf{Z}
%\\ \,\, \mathbf{1}^T
\end{bmatrix}= \begin{bmatrix}
\mathbf{P} \\
\mathbf{X}
%\\ \,\, \mathbf{1}^T
\end{bmatrix} - \begin{bmatrix}
\mathbf{E}_P \\
\mathbf{E}_X
%\\ \mathbf{0}^T
\end{bmatrix}, \nonumber\\
&\,\; \mathbf{1}^T_c \mathbf{Y} = \mathbf{1}^T_N, \,\, \mathbf{Y} \in \mathbb{R}_+^{c \times N}, \label{cond:probsim2}\\
&\, y_{i,j} = 0 \,\, \mathrm{if}  \,\, i \notin L_j, \,\, i = 1, 2,  \dots, c-1, \,\, \forall j, \label{cond:al2}\\
&\sum_{j \in G_k}  \sum_{i=1}^{c-1} y_{i,j}  \geq 1 \,\, \mathrm{if} \,\,  \mathop{\cup}_{j \in G_k} L_j \neq \{c\}, \forall k, \label{cond:nonnull} \\
& \sum_{j \in G_k} y_{i,j}  \leq 1, \,\, i = 1, 2, \dots, c-1, \,\,  \forall k. \label{cond:unique1}
\end{align}
Constraints (\ref{cond:probsim2}) and (\ref{cond:al2}) are inherited from the original formulation. The constraint in (\ref{cond:nonnull}), assumes that there is at least one non-null identity in a photo unless all the instances in a photo are explicitly labeled as null. This constraint is enforced to avoid the trivial solution that all the instances are treated as belonging to the null class.
 A similar constraint has been considered by \cite{Luo2010} and \cite{Zeng2013} via restricting the candidate labeling set and confining the feasible space of PPM, respectively. The constraint in (\ref{cond:unique1}) enforces the uniqueness of non-null identities. Note that this framework can be easily tailored to handle other prior knowledge (e.g. must/cannot-link constraints, prior statistics) by regularizing the labeling matrix. This problem can be solved by following the similar relaxation procedures for solving (\ref{eqn:Y_relaxed1}). The optimization procedure is summarized in Algorithm \ref{alg:alm2}.

Following the relaxation procedure in Section \ref{MCar_sec:opt}, we can reformulate (\ref{eqn:HardExt}) as
\begin{align}
\min_{\mathbf{H},\mathbf{E}, \mathbf{\Lambda}, \mu} \,\,&   \| \mathbf{H} \|_*  + \lambda \|\mathbf{E}_X\|_1  +  \gamma \|\mathbf{P} -\mathbf{E}_P\|_1 \nonumber \\
& + \left \langle  \mathbf{\Lambda} ,
\mathbf{H}_{obs} -\mathbf{H}
%\\ \,\, \mathbf{1}^T
-
\mathbf{E}
 \right \rangle  + \frac{\mu}{2} \left\|
\mathbf{H}_{obs} - \mathbf{H}
 -
\mathbf{E}
 \right\|_F^2, \label{eqn:Y_relaxed_augR2} \\
\text{s.t.}
&\,\; \mathbf{1}^T_c \mathbf{Y} = \mathbf{1}^T_N, \,\, \mathbf{Y} \in \mathbb{R}_+^{c \times N}, \label{cond:probsim22}\\
&\, y_{i,j} = 0 \,\, \mathrm{if}  \,\, i \notin L_j, \,\, i = 1, 2,  \dots, c-1, \,\, \forall j, \label{cond:al22}\\
&\sum_{j \in G_k}  \sum_{i=1}^{c-1} y_{i,j}  \geq 1 \,\, \mathrm{if} \,\,  \mathop{\cup}_{j \in G_k} L_j \neq \{c\}, \forall k, \label{cond:nonnull2} \\
& \sum_{j \in G_k} y_{i,j}  \leq 1, \,\, i = 1, 2, \dots, c-1, \,\,  \forall k. \label{cond:unique2}
\end{align}
We use a similar procedure of Algorithm \ref{alg:alm_W} presented in Section \ref{MCar_sec:opt} to solve (\ref{eqn:Y_relaxed_augR2}). We again use the projection method to guide the process of matrix completion such that the constraints on $\mathbf{Y}$ are satisfied. Additionally, the projection of $\mathbf{Y}$ handles the group constraints such that the labeling constraints between instances are satisfied. Hence, we project $\mathbf{Y}$ to the feasible regions indicated by (\ref{cond:al22}), (\ref{cond:nonnull2}), and (\ref{cond:unique2}) one at a time, and each one is followed by the projection onto the feasible region indicated by (\ref{cond:probsim22}) to ensure that each column of $\mathbf{Y}$ lies in the probability simplex.
The detailed procedure is summarized in Algorithm \ref{alg:alm2}. This algorithm can be easily extended to handle ambiguously labeled data with labeling imbalance by taking $\bar{\mathbf{H}}$ as input with proper manipulation on the projection steps of $\bar{\mathbf{Y}}$.% [WLOG, we can also incorporate the ICE]

 \begin{algorithm}[htp!]
  \caption{The optimization algorithm for (\ref{eqn:Y_relaxed_augR2})}
  \begin{algorithmic}[1]
    \Require
      $\mathbf{P} \in \mathbb{R}^{c \times N}$, $\mathbf{X} \in \mathbb{R}^{m \times N}$, $L_j \,\, \forall j$, $G_k \,\, \forall k$,  $\lambda$, and $\gamma$.\\
    \textbf{Initialization:} $\mathbf{Y}=\mathbf{0}$, $\mathbf{Z} = \mathbf{0}$, $\mu > 0, \mu_{\max} > 0$,  $\rho > 1$, $\mathbf{\Lambda} = [\mathbf{\Lambda}_P; \mathbf{\Lambda}_X] = \mathbf{H}_{obs} / \|\mathbf{H}_{obs}\|_2$;
    \While {not converged}
      \State $\mathbf{E}_P = \mathbf{P} - \mathcal{S}_{\gamma \mu^{-1}}[ \mathbf{Y} - \mu^{-1} \mathbf{\Lambda}_P]$;
      \State $\mathbf{E}_X = \mathcal{S}_{\lambda \mu^{-1}}[ \mathbf{X} - \mathbf{Z} + \mu^{-1} \mathbf{\Lambda}_X]$;
      \State $(\mathbf{U}, \mathbf{\Sigma}, \mathbf{V}) = \mathrm{svd}\left(\mathbf{H}_{obs} - \mathbf{E} + \mu^{-1} \mathbf{\Lambda}\right)$;
      \State $\mathbf{H} = \mathbf{U} \mathcal{S}_{\mu^{-1}}[\mathbf{\Sigma}] \mathbf{V}^T $;
      \State $\mathbf{\Lambda} = \mathbf{\Lambda} + \mu \left(\mathbf{H}_{obs} - \mathbf{H} - \mathbf{E} \right)$;
      \State $\mu = \min(\rho \mu , \mu_{\max} )$;
      \State \textbf{Project }$\mathbf{Y}$:
      \LineComment{Line: 11-13: Projection for (\ref{cond:al22}) and (\ref{cond:probsim22}) }
      \State $\mathbf{Y} = \max(\mathbf{Y}, 0)$; \label{line:p2_1}  % $ y_{i,j} \in \mathbb{R}_+$;
      \State $y_{i,j} = 0 \,\, \mathrm{if}  \,\, i \notin L_j,\,\, i = 1, 2, \dots, c-1,\forall j$ \label{line:p2_2};
      \State $\mathbf{y}_j = \mathbf{y}_j / \|\mathbf{y}_j\|_1,$ $\forall j$;% $\mathbf{1}^T_c \mathbf{Y} = \mathbf{1}^T_N$;
      \LineComment{Line: 15-22: Projection for (\ref{cond:nonnull2}) and (\ref{cond:probsim22})}
      \For {$k=1:K$}
      \If {$\cup_{j \in G_k} L_j \neq \{c\}$}
      \For {$i=1:c-1, j \in G_k$}
\\   \quad\quad\quad\quad\quad\,\,\,\,   $ y_{i,j} =  y_{i,j} / \min( \sum_{g \in G_k}  \sum_{i=1}^{c-1} y_{i,g} , 1 )$;
      \EndFor
      \EndIf
      \EndFor
      \State $\mathbf{y}_j = \mathbf{y}_j / \|\mathbf{y}_j\|_1,$ $\forall j$;% $\mathbf{1}^T_c \mathbf{Y} = \mathbf{1}^T_N$;
      \LineComment{Line: 24-29: Projection for (\ref{cond:unique2}) and (\ref{cond:probsim22})}
      \For {$k=1:K$}
      \For {$i=1:c-1, j \in G_k$}
      \State $ y_{i,j} =  y_{i,j} / \max( \sum_{g \in G_k} y_{i,g}, 1 )$;
      \EndFor
      \EndFor
      \State $\mathbf{y}_j = \mathbf{y}_j / \|\mathbf{y}_j\|_1,$ $\forall j$;% $\mathbf{1}^T_c \mathbf{Y} = \mathbf{1}^T_N$;
    \EndWhile
    \Ensure
        $(\mathbf{H}, \mathbf{E})$
  \end{algorithmic}\label{alg:alm2}
\end{algorithm}

\section{Experimental Results} \label{Chapter_MCar:sec:results}

We use the Labeled Faces in the Wild (LFW) dataset \cite{Huang2007} and the CMU PIE dataset with synthesized ambiguous labels to evaluate the performance of our method under various controlled parameter settings. Furthermore, we use the \emph{Lost} dataset \cite{Cour2009} and the Labeled Yahoo! News dataset \cite{Berg2004,Guillaumin2010} to demonstrate the effectiveness of our method in real-world applications. For the LFW, CMU PIE, and \emph{Lost} datasets, we use face images in gray scale of range $[0, 1.0]$. Each instance is preprocessed with histogram equalization and converted into a column feature vector.

\subsection{Parameters} \label{sec:exp:parameters}

It is interesting to observe that (\ref{eqn:Hard}) becomes asymptotically similar to the formulation of Robust Principle Component Analysis (RPCA) \cite{Candes2011} when the dimension of the data feature is much larger than the number of classes. Motivated by this fact, we fix $\lambda$ as
\begin{align}
     \lambda_o = \frac{1}{\sqrt{\max(c+m, N)}}, \label{eqn:lambda_opt}
\end{align}
which is the tradeoff parameter suggested in RPCA. $\gamma$ is a tuning parameter that controls the sparsity of the soft labeling vectors. For MCar-based methods, we use $\gamma = 2 \lambda_0$ to encourage stronger sparsity of the labeling vector than that of feature noise. For ICE, we set the elimination factor $f_e$ as 0.5, and set the maximum number of iterations as 5. These parameters yield good results in general, and we will investigate the sensitivity of parameters in Section \ref{subsec:parameter_sen}.

\begin{figure*}
        \centering
        \begin{subfigure}[b]{0.32\textwidth}
                \includegraphics[trim=10 3 18 19,clip,width=\textwidth]{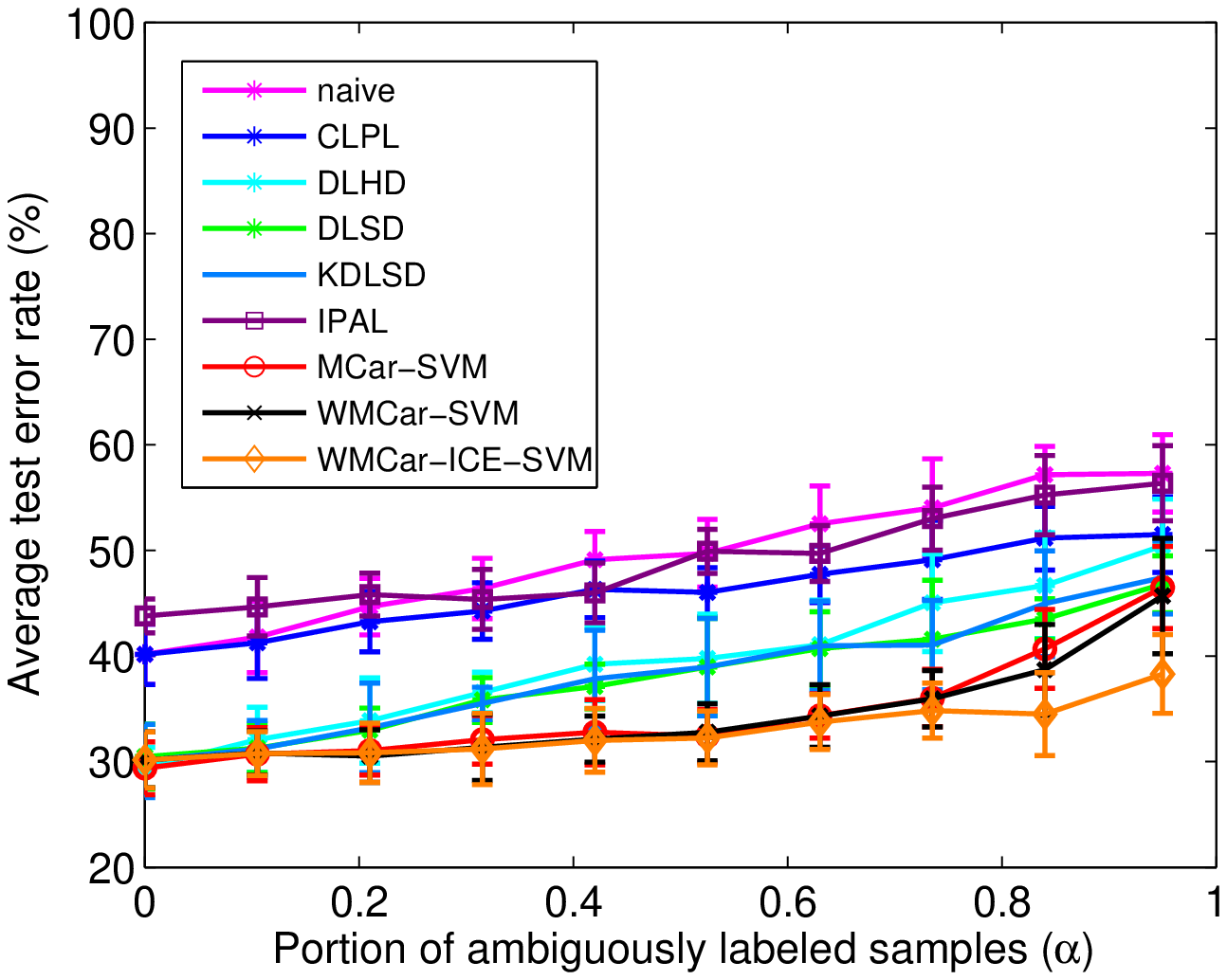}
                %\caption{$\alpha= 0.0 : 0.95$, $\beta =2 $ for FIW10b, inductive}
                \caption{}\label{fig:fiw10b_a}
        \end{subfigure}\,
        %~ %add desired spacing between images, e. g. ~, \quad, \qquad etc.
        \begin{subfigure}[b]{0.32\textwidth}
                \includegraphics[trim=10 3 18 19,clip,width=\textwidth]{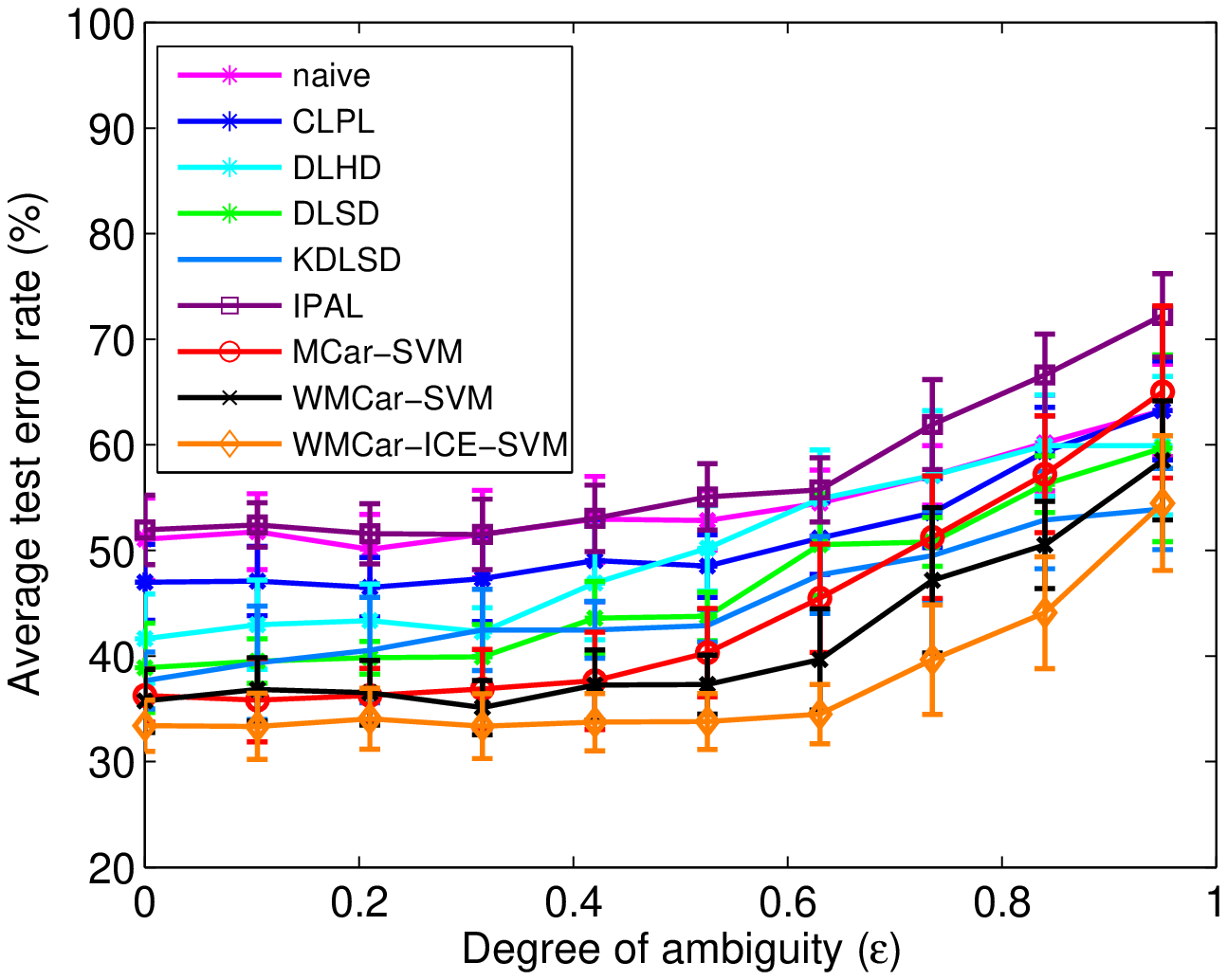}
                %\caption{$\alpha= 1.0$, $\beta =1$, $\epsilon \in [1/(c-1), 1]$ for FIW10b, inductive}
                %\label{fig:R0_far}
                \caption{}\label{fig:fiw10b_b}
        \end{subfigure}\,
        %add desired spacing between images, e. g. ~, \quad, \qquad etc.
       \begin{subfigure}[b]{0.32\textwidth}
                \includegraphics[trim=10 3 18 19,clip,width=\textwidth]{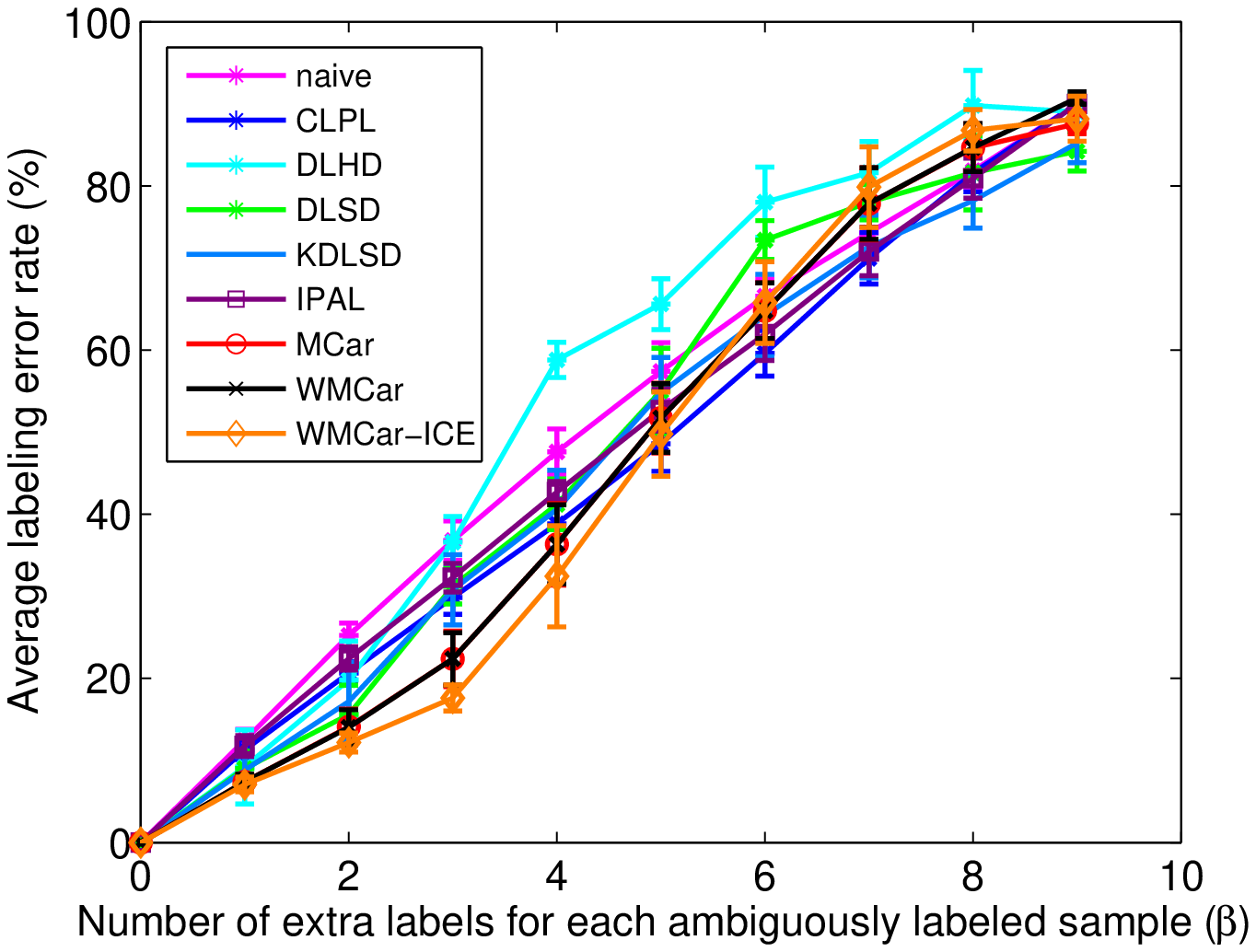}
                %\caption{$\alpha = 1$, $\beta = 0:9$ for FIW10b, \textcolor[rgb]{1.00,0.00,0.00}{transductive or inductive (run exp for CLPL)}}
                %\label{fig:initialL0}
                \caption{}\label{fig:fiw10b_c}
        \end{subfigure}\,
        \caption{Performance comparisons on the FIW(10b) dataset. (a) $\alpha \in [0, 0.95]$, $\beta=2$, \emph{inductive} experiment. (b) $\alpha= 1.0$, $\beta =1$, $\epsilon \in [1/(c-1), 1]$, \emph{inductive} experiment. (c) $\alpha= 1.0$, $\beta  \in [0,1, \dots, 9]$, \emph{transductive} experiment.}
        \label{fig:lfw_set}
\end{figure*}

\subsection{Experiments with the Synthesized Dataset}
We conduct two types of controlled experiments suggested in \cite{Cour2011}.  For the \emph{inductive} experiment, the dataset is evenly split into ambiguously labeled training set and unlabeled testing set. The proposed methods, MCar/WMCar-SVM and WMCar-ICE-SVM, learn a multi-class linear SVM \cite{CC01a} with the disambiguated labels provided by MCar/WMCar and WMCar-ICE, respectively. The testing data is then classified using the learned classifier.
For the \emph{transductive} experiment, all the data is used as the ambiguously labeled training set.

We follow the ambiguity model defined in \cite{Cour2011} to generate ambiguous labels in the controlled experiment. Note that $\alpha$ denotes the number of extra labels for each instance, and $\beta$ represents the portion of the ambiguously labeled data among all instances. The degree of ambiguity $\epsilon$ indicates the maximum probability that an extra label co-occurs with a true label, over all labels and instances. Each controlled experiment is repeated $20$ times.
We report the average testing (labeling) error rate for inductive (transductive) experiment, where the testing (labeling) error rate is the ratio of the number of erroneously labeled instances to the total number of instances in the testing (training) set. The standard deviations are plotted as error bars in the figures.

We compare the proposed MCar-based methods with several state-of-the-art ambiguous learning approaches for single instances with ambiguous labeling: CLPL \cite{Cour2011}, DLHD/DLSD \cite{Chen2013}, KDLSD \cite{Chen2014all}, and IPAL \cite{Zhang2015stp}. We report the performance of these methods when the experimental results are available in their papers. Otherwise, we use the configuration suggested in their papers to conduct the experiments. We use `naive' \cite{Cour2011} as the baseline method, which learns a classifier from minimizing the trivial $0/1$ loss.

\subsubsection{The LFW Dataset}

The FIW(10b) dataset \cite{Cour2011} consists of the top $10$ most frequent subjects selected from the LFW dataset \cite{Huang2007}, and the first $50$ face images of each subject are used for evaluation.
We use the cropped and resized face images readily provided by the authors of \cite{Cour2011}, where the face images are of $45 \times 55$ pixels.

Figures \ref{fig:fiw10b_a} and \ref{fig:fiw10b_b} show the results of the inductive experiments. Figure \ref{fig:fiw10b_a} shows that the MCar-based methods significantly outperform all the other methods when the portion of ambiguously labeled data is larger than 0.2. The performance of WMCar is comparable to that of MCar since the ambiguously labeled data generated by this ambiguity model does not substantially result in labeling imbalance. WMCar-ICE demonstrates better performance than MCar and WMCar when more than 0.7 portions of the instances are ambiguously labeled.
An explanation is that ICE eliminates the candidates based on the ordering of the least soft labeling score of each instance. This prioritization step can effectively benefit from a large portion of ambiguously labeled samples (i.e., large $\alpha$) that usually carries a diverse aspect of soft labeling scores. When the portion of ambiguously labeled samples is small, the improvement due to ICE becomes insignificant.

Figure \ref{fig:fiw10b_b} shows that MCar outperforms prior methods over various degrees of ambiguity except when $\epsilon > 0.7$. Thus, MCar yields improved performance at low and intermediate levels of ambiguity, but it becomes susceptible at high levels of ambiguity.
One explanation is that both the true label and the extra labels of a subject will result in low-rank component of the labeling matrix when they are likely to co-occur in high degree of ambiguity. Consequently, separating the true label from the extra labels in MCar becomes challenging. Another explanation is that a high degree of ambiguity results in labeling imbalance, which causes the performance degradation of MCar. To verify this, we obtain the label distribution by counting the number of label occurrences in the candidate labeling sets for each class. We define the imbalance factor as the ratio of the maximum to the minimum value in the label distribution. The average imbalance factor varies from 1.33 to 3.58 as the degree of ambiguity increases. This confirms that WMCar outperforms MCar in high degree of ambiguity since WMCar is effective in mitigating the impact of labeling imbalance. Furthermore, WMCar-ICE outperforms WMCar by iteratively removing the least likely candidate labels from the candidate labeling sets. This experiment demonstrates that the labeling imbalance can cause performance degradation even though there is no class imbalance among the number of groundtruth faces per class.

In Figure \ref{fig:fiw10b_c}, MCar-based methods outperform other approaches only when the number of extra labels is less than $5$ in the transductive experiment. This shows that MCar-based methods cannot be effective when the labeling is severely cluttered such that the low-rank approximation of heterogeneous feature fails. Similar to the controlled parameter setting in Figure \ref{fig:fiw10b_a}, the ambiguously labeled data generated by this ambiguity model does not substantially result in labeling imbalance. Hence, the performance of WMCar is comparable to that of MCar, and WMCar-ICE slightly outperforms WMCar.

Figure \ref{fig:X_decomp_fiw} shows the intermediate results of low-rank decomposition of the feature matrix using MCar. Note that variations due to illumination, occlusions (e.g. eyeglasses, hand), and expressions are suppressed such that the low-rank component of a subject is preserved. In contrast to MCar-based methods, the discriminative methods (e.g. naive, CLPL) and IPAL are susceptible to such variations. Furthermore, it also demonstrates the robustness of our methods even though the face images are not perfectly aligned. The proposed method outperforms the dictionary-based methods \cite{Chen2013,Chen2014all} for all cases except when there is severe ambiguity. Note the low-rank approximation of MCar operates on the feature matrix and ambiguous labeling matrix as a whole by concatenating them such that the actual labels and the low-rank component of feature matrix are recovered simultaneously. This essentially demonstrates the advantage of the proposed method over the DLHD/DLSD and KDLSD methods that iteratively alternate between confidence and dictionary update.

\subsubsection{The CMU PIE Dataset}

The CMU PIE dataset contains face images from $68$ subjects of different poses, illumination conditions, and expressions.
Following the protocol presented in \cite{Chen2013}, we select the $18$ subjects for evaluation. Each subject has 21 images under different illumination conditions, and the face images are resized to $40 \times 48$ pixels.

\begin{figure*}
        \centering
        \begin{subfigure}[b]{0.4\textwidth}
                \includegraphics[width=\textwidth]{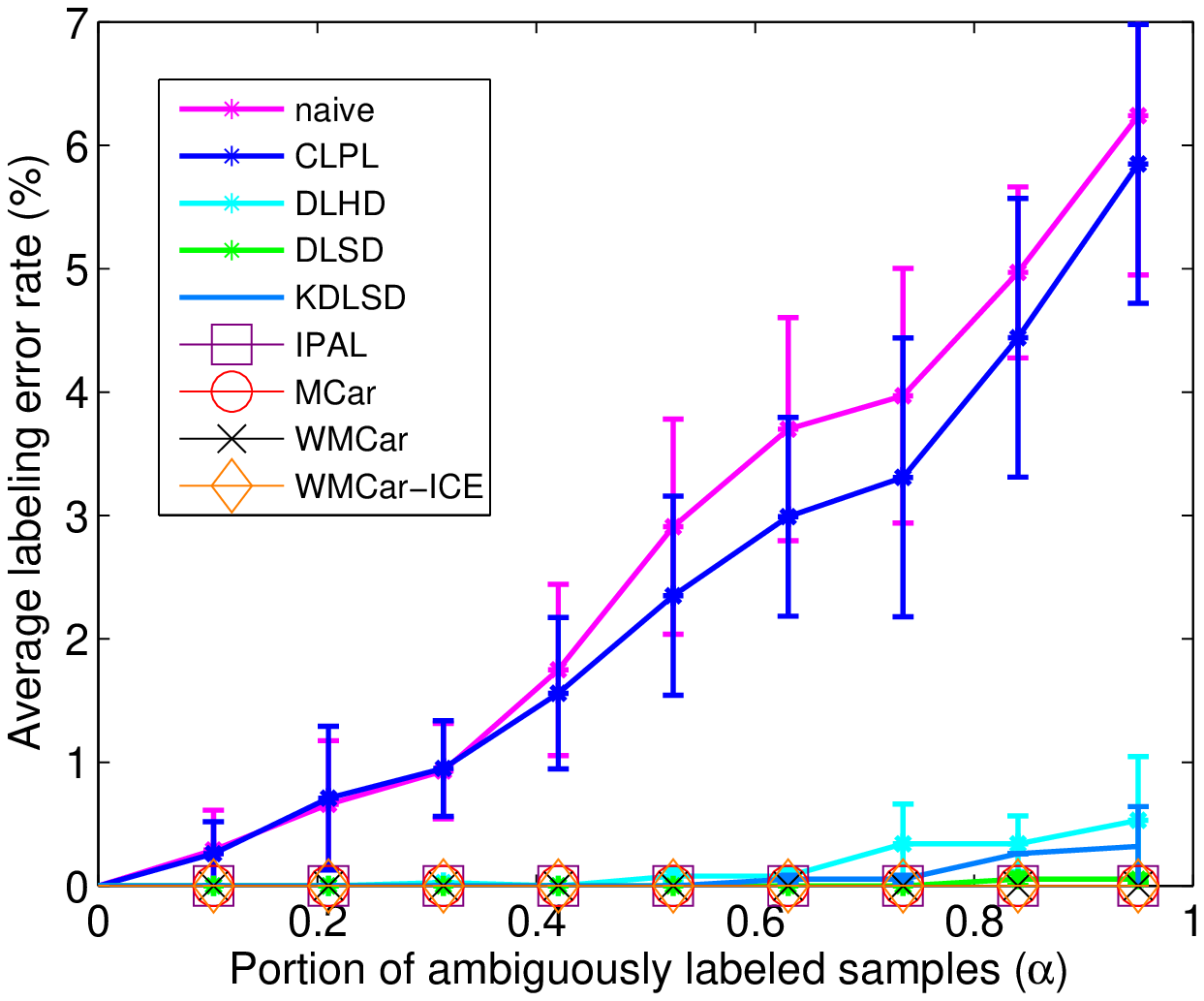}
                \caption{}\label{fig:pie_a}
        \end{subfigure}\,
        \begin{subfigure}[b]{0.4\textwidth}
                \includegraphics[width=\textwidth]{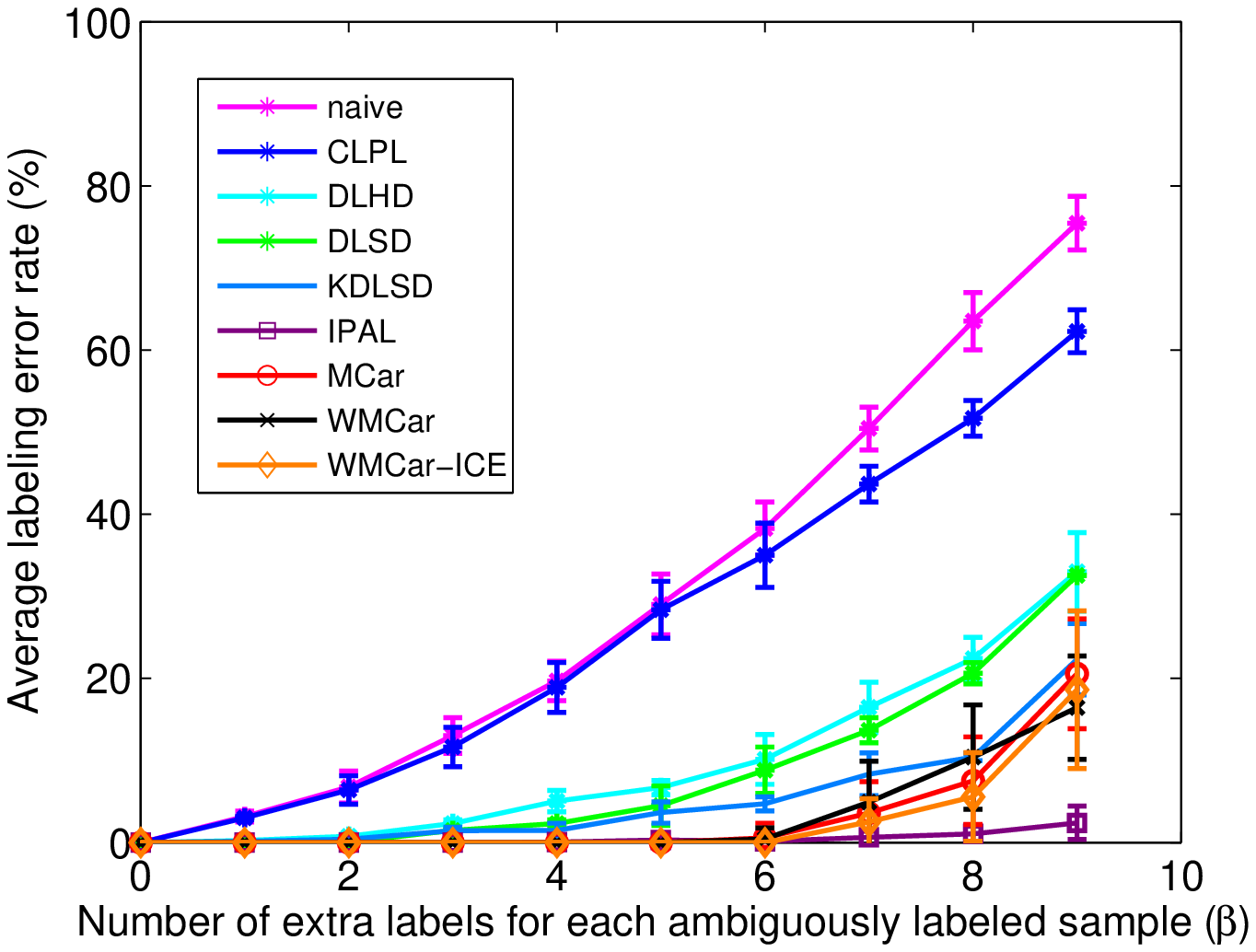}
                \caption{}\label{fig:pie_b}
        \end{subfigure}\,
        \caption{Performance comparisons on the CMU PIE dataset. (a) $\alpha \in [0, 0.95]$, $\beta=2$, \emph{transductive} experiment. (b) $\alpha= 1.0$, $\beta  \in [0,1, \dots, 9]$, \emph{transductive} experiment.}\label{fig:pie_set}
\end{figure*}

\begin{figure}
        \centering
        \begin{subfigure}{\columnwidth}
        \centering
                \includegraphics[trim= 60 40 45 20,clip,width=0.9\textwidth]{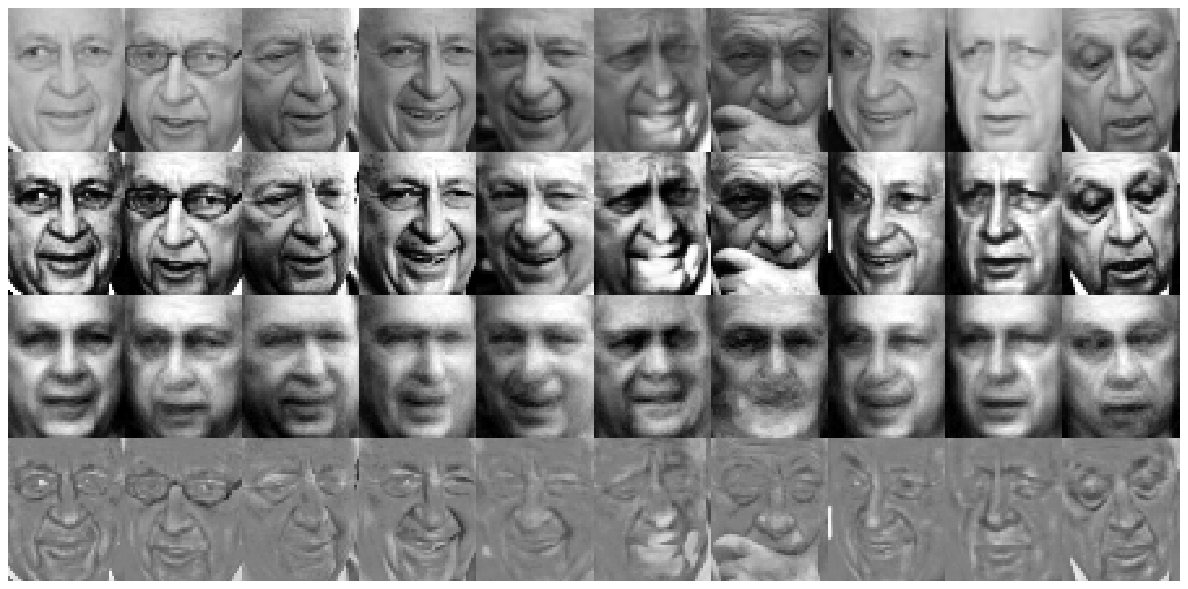}
                \caption{}\label{fig:X_decomp_fiw}
        \end{subfigure}\\
        %~ %add desired spacing between images, e. g. ~, \quad, \qquad etc.
        \begin{subfigure}{\columnwidth}
        \centering
                \includegraphics[trim= 60 40 48 18,clip,width=0.9\textwidth]{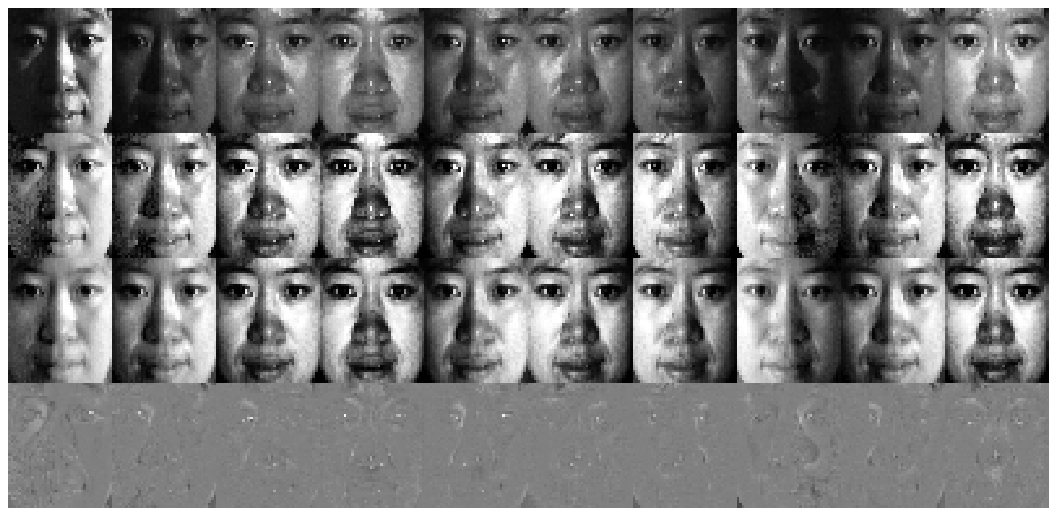}
                \caption{}\label{fig:X_decomp_pie}
        \end{subfigure}
        \caption{Subsets of images from (a) FIW(10b) and (b) CMU PIE dataset demonstrate the low-rank decomposition of feature matrix in MCar: the original face images, histogram-equalized images $\mathbf{X}$, low-rank component $\mathbf{Z}$, and noisy component $\mathbf{E}_X$, from the first row to the forth row, respectively.}
\end{figure}

We synthesize the ambiguous labels based on the controlled parameters. The results of two transductive experiments for CMU PIE dataset are shown in Figures \ref{fig:pie_a} and \ref{fig:pie_b}. In Figure \ref{fig:pie_a}, MCar-based methods and IPAL recover all the label ambiguity for various portions of ambiguously labeled samples. In Figure \ref{fig:pie_b}, our proposed methods consistently outperform most of the state-of-the-art methods except IPAL as we increase the number of extra labels for each ambiguously labeled sample. Since the CMU PIE dataset is collected in a constrained environment, the collective face images of a subject are well-modeled by low-rank approximation. Hence, MCar demonstrates marginally improvements over most of the methods in this dataset. This can be seen by visualizing the intermediate results of low-rank decomposition of the feature matrix using MCar as shown in Figure \ref{fig:X_decomp_pie}. Besides, the IPAL method outperforms our methods when $\beta > 6$. Since the IPAL method utilizes the locally linear embedding for label propagation, which is effective in learning the underlying structure of data that has plenty of samples collected in the constrained environment. Hence, IPAL is able to recover the severely cluttered labels that MCar-based methods fail to approximate it as a low-rank matrix.

\subsection{Experiments with Real-world Dataset}
We conduct experiments on the \emph{Lost} dataset and Labeled Yahoo! News dataset where the ambiguous labeling are collected in the real world. In the Labeled Yahoo! News dataset, we consider the labeling constraints between instances.

\subsubsection{The Lost Dataset}
\begin{figure}
\centering
\includegraphics[trim= 30 10 50 30,clip,width=0.5\textwidth]{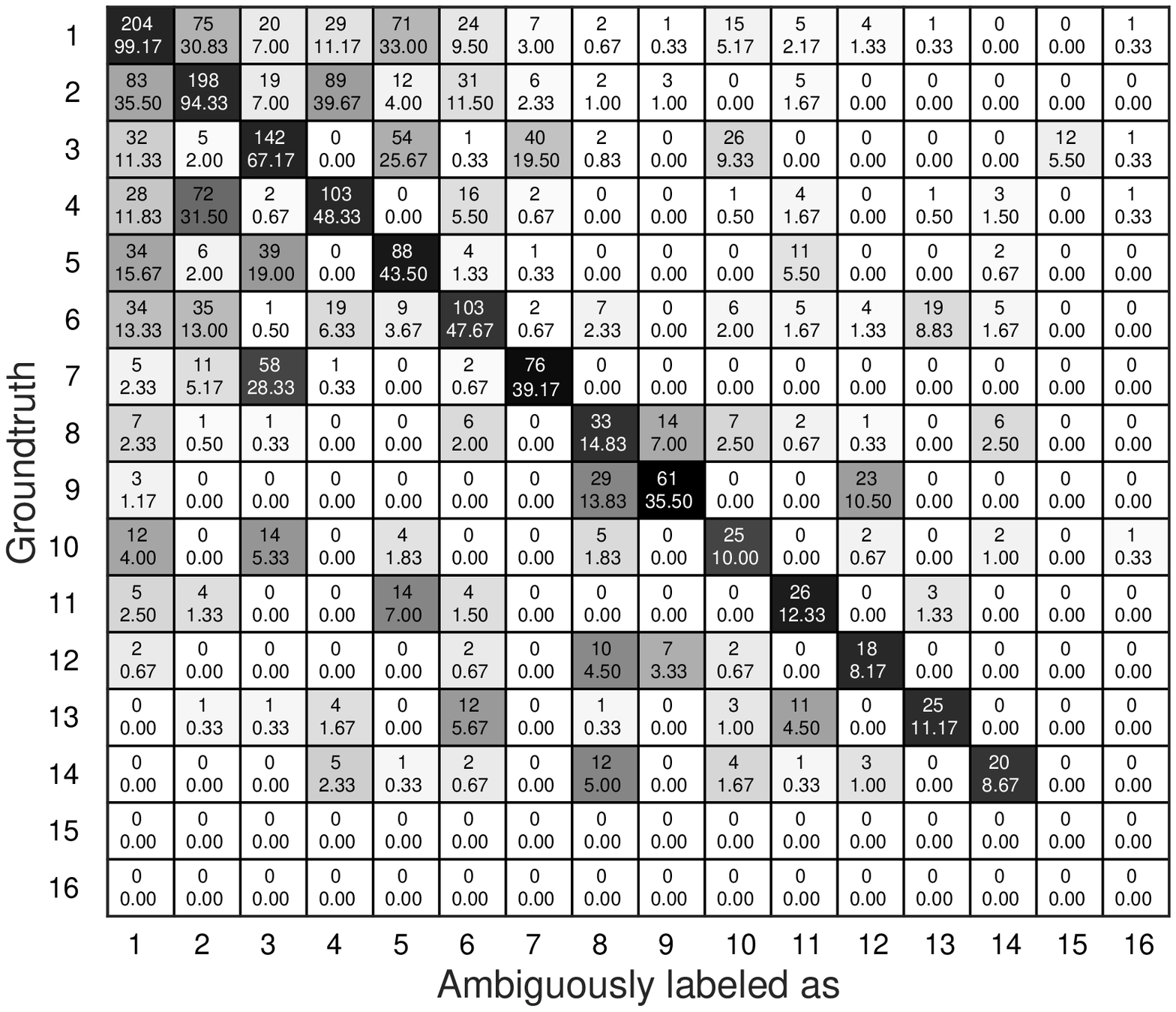}
\caption{The confusion matrix of the ambiguous labeling in \emph{Lost} $(16,8)$ dataset. The upper number of each square accounts for the number of occurrences of a candidate label, whereas the lower number of each square is computed by accumulating the soft labeling score of each occurrence of a candidate label.}\label{fig:label_dist_lost}
\end{figure}

The \emph{Lost} dataset consists of face images and ambiguous labels automatically extracted using the screenplays provided in the TV series \emph{Lost}. We use the \emph{Lost} $(16,8)$ dataset released by the authors of \cite{Cour2009} for evaluation. The \emph{Lost} $(16,8)$ dataset consists of 1,122 registered face images from $8$ episodes, and the size of each is $60 \times 90$ pixels. The labels cover $16$ subjects, but only $14$ of them appear in the dataset. Figure \ref{fig:label_dist_lost} illustrates the confusion matrix of ambiguous labeling, which exhibits labeling imbalance.

We compare MCar-based methods with `naive', CLPL , MMS \cite{Luo2010}, and IPAL \cite{Zhang2015stp}. No labeling constraint between instances is considered in this experiment. Results are shown in Table \ref{tab:lostcomp}.  It can be seen from this table that MCar-based methods significantly outperform CLPL and MMS. This shows that MCar-based methods resolve the ambiguity and handles variations of instances in the TV series much better when compared to discriminative methods. Note that the performance of MMS is close to that of CLPL since the ambiguous loss functions of both methods become similar when no labeling constraint between the instances is considered.

\begin{figure}
\centering
\includegraphics[width=0.45\textwidth]{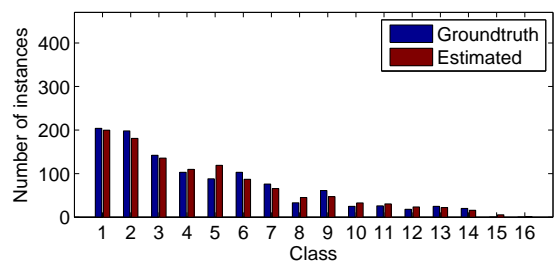}
\caption{The groundtruth label distribution of the \emph{Lost} $(16,8)$ dataset. `Groundtruth' denotes the number of instances per class counted from the groundtruth labels, and  `Estimated' denotes the estimate of the groundtruth label distribution from the ambiguous labels.}\label{fig:data_imbalance_stat}
\end{figure}

Figure \ref{fig:data_imbalance_stat} demonstrates that the groundtruth label distribution estimated by (\ref{eqn:est_sub_dist}) is close to the groundtruth. Hence, WMCar can effectively utilize this information to compensate the labeling imbalance. Although WMCar slightly outperforms MCar, the approach that combines WMCar and ICE (WMCar-ICE) significantly outperforms MCar. On the other hand, MCar-ICE is inferior to WMCar-ICE since the ICE procedure can inadvertently remove the candidates corresponding to minor labels without considering the labeling imbalance.
It is challenging for IPAL to exploit the underlying structure of scarcity labeled data and deal with labeling imbalance. Hence, IPAL cannot successfully resolve the label ambiguity in this dataset. We tailor the RPCA \cite{Lin2009} and MC-Pos \cite{Cabral2014} to solve the ambiguous learning problem by trivially taking the heterogeneous matrix as input and predicting the labels from the soft labeling matrix of output with (\ref{eqn:ypredict}). The experimental result shows that existing low-rank approximation methods cannot substantially resolve the label ambiguity.

\begin{table}
\begin{center}
\begin{tabular}{lr}
  \toprule[1.0pt]
  \head{Method\quad\quad\quad\quad\quad\quad\quad\quad\quad}  & \head{Error Rate}\\
  \midrule
  naive  & 18.6 \%\\
  CLPL  \cite{Cour2011}  & 12.6 \%\\
  MMS \cite{Luo2010}  & 11.4 \%\\
  IPAL \cite{Zhang2015stp}  & 22.9 \%\\
  \midrule
  RPCA \cite{Lin2009}  & 29.9 \% \\
  MC-Pos \cite{Cabral2014}  & 23.6 \% \\
  \midrule
  MCar   & 8.5 \%\\
  WMCar   & 8.2 \%\\
  MCar-ICE   & 8.0 \%\\
  WMCar-ICE   & 5.2 \% \\
  \bottomrule[1.0pt]
\end{tabular}
\end{center}
\caption{Labeling error rates for the \emph{Lost} $(16,8)$ dataset (available at http://www.timotheecour.com/tv\_data/tv\_data.html).}
\label{tab:lostcomp}
\end{table}

\begin{table}
\begin{center}
\begin{tabular}{lr}
  \toprule[1.0pt]
  \head{Method\quad\quad\quad\quad\quad\quad\quad \,\,}  & \head{Error Rate}\\
  \midrule
  CL-SVM    & 23.1 \% $\pm$ 0.6 \%\\
  MIMLSVM \cite{Zhou2006}  & 25.3  \% $\pm$ 0.3 \%\\
  MM-SVM \cite{Berg2004} & 21.9 \% $\pm$ 1.0 \%\\  
  MMS \cite{Luo2010}  &14.3 \% $\pm$ 0.5 \% \\
  LR-SVM \cite{Zeng2013}  & 19.2 \%  $\pm$ 0.4 \%\\
  \midrule
  MCar-SVM   &  14.5 \%  $\pm$ 0.4 \%\\
  WMCar-SVM   & 13.6 \% $\pm$ 0.8 \%\\
  MCar-ICE-SVM   & 15.0 \%  $\pm$ 1.0 \% \\
  WMCar-ICE-SVM   &  12.9 \% $\pm$ 0.8 \%\\
  \bottomrule[1.0pt]
\end{tabular}
\end{center}
\caption{Average testing error rates for the Labeled Yahoo! News dataset (available at http://lear.inrialpes.fr/data).}
\label{tab:yahoo}
\end{table}

\subsubsection{The Labeled Yahoo! News Dataset}
The Labeled Yahoo! News dataset contains fully annotated faces in images with names in the captions. It consists of 31,147 detected faces from 20,071 images. We use the precomputed SIFT feature of dimension 4,992 extracted from that face images provided by Guillaumin \emph{et al.} \cite{Guillaumin2010}. Following the protocol suggested in \cite{Luo2010}, we retain 214 subjects with at least twenty occurrences in the captions. The remaining face images and names are treated as belonging to the additional null class. The ambiguous labeling is unbalanced in this dataset, where the number of labels present in the captions ranges from 20 to 1,917 with mean and standard deviations equal to 64.6 and 147.3, respectively. The top two subjects that are present most frequently in the captions are `george\_w\_bush' and `saddam\_hussein'. We conduct experiments on five training/testing splits by randomly selecting $80\%$ of images and their associated captions as training set, and the rest are used as testing set. In each split, we also maintain the ratio between the number of training and testing instances from each subject.

The baseline approaches are CL-SVM, MIMLSVM \cite{Zhou2006}, and MM-SVM \cite{Berg2004}. The implementation details of CL-SVM and MIMLSVM are provided in \cite{Luo2010}. We implemented the maximal likelihood assignment (MM) \cite{Berg2004} to resolve the label ambiguity in the training data and train a multi-class linear SVM \cite{CC01a} to classify the testing data.
To ensure a fair comparison, the language modeling of news captions in \cite{Berg2004} is not included in the implementation.
 We compare with two state-of-the-art ambiguous labeling methods that consider labeling constraints between instances: MMS \cite{Luo2010} and LR-SVM \cite{Zeng2013}, which are based on discriminative model and low-rank framework, respectively. We resolve the label ambiguities in the training set using (\ref{eqn:HardExt}) and train a multi-class linear SVM \cite{CC01a} to classify the testing data. Our MCar-SVM algorithm exhibits a slightly $0.2\%$ higher error rate as compared to MMS.
An explanation is that MCar relying on the low-rank approximation for ambiguity resolution is particularly sensitive to labeling imbalance. This results in performance degradation in the learned classifier since the output labels of MCar are potentially biased toward the majority labels.

Compared to the LR-SVM method, the MCar-SVM algorithm demonstrates $4.7\%$ improvement on the testing accuracy. Since MCar assigns the labels across all instances via low-rank approximation of heterogeneous feature matrix, it is more effective than the LR-SVM method, which updates the PPM and the low-rank subspace of each class alternately.
When we consider the labeling imbalance and utilize the ICE procedure, our proposed WMCar-ICE-SVM outperforms MCar-SVM by $1.6\%$

\begin{figure}
\centering
\includegraphics[width=0.45\textwidth]{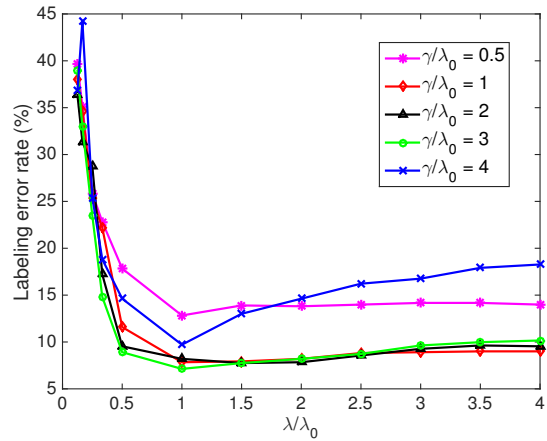}
\caption{Labeling error rates of WMCar evaluated with a set of parameters $(\lambda, \gamma)$ in the \emph{Lost} $(16,8)$ dataset.
}
\label{fig:tune_lambda_gamma}
\end{figure}

\subsection{Sensitivity of Parameters} \label{subsec:parameter_sen}

We use the \emph{Lost} $(16,8)$ dataset to conduct the sensitivity analysis of MCar-based methods. In Figure \ref{fig:tune_lambda_gamma}, we evaluate the performance of WMCar over a set of parameters $(\lambda, \gamma)$. We observe that the labeling error rate is relatively low when $\lambda$ approaches $\lambda_0$ with respect to various $\gamma$. Hence, we conclude that the tradeoff parameter suggested in RPCA is applicable or at least suggests a reasonable value for $\lambda$.

\begin{figure}
\centering
\includegraphics[width=0.45\textwidth]{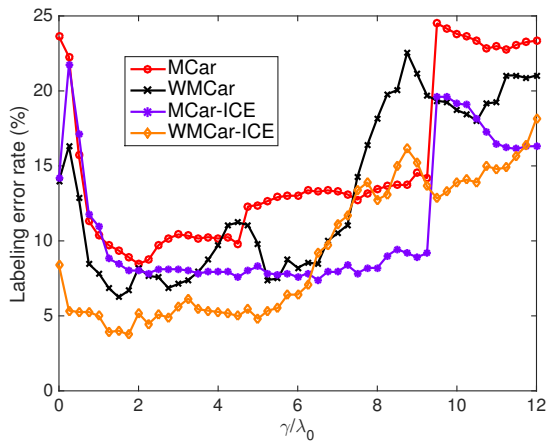}
\caption{Labeling error rates of MCar-based methods versus $\gamma$ in the \emph{Lost} $(16,8)$ dataset with $\lambda = \lambda_o$.}\label{fig:parameter_tune}
\end{figure}

In Figure \ref{fig:parameter_tune}, we evaluate the performance of MCar-based methods with various $\gamma$ and a fixed $\lambda = \lambda_o$. For ICE, we set the elimination factor $f_e$ as 0.5, and we set the maximum number of iterations as 5. Note that $\gamma$ controls the sparsity of the soft labeling matrix $\mathbf{Y}$, and a larger $\gamma$ will encourage a stronger sparsity on $\mathbf{Y}$. For $\gamma \in [1, 4]\lambda_o$, a moderate sparsity of $\mathbf{Y}$ is helpful in predicting the actual label from the ambiguous labels. As $\gamma$ (sparsity of $\mathbf{Y}$) significantly increases, the performance degrades. One explanation is that a very strong sparsity on $\mathbf{Y}$ imposes an immediately hard decision on the labels, which may have an adverse effect on low-rank approximation. On the other hand, a mild sparsity on $\mathbf{Y}$ allows an unsure face image to be represented by the feature subspaces of those individuals that resemble the unsure face image.
The selected parameter $\gamma = 2\lambda_o$ yields good performance for the MCar-based methods as illustrated in Figure \ref{fig:parameter_tune}.
We observe that WMCar-ICE is less sensitive to $\gamma$ since the ICE procedure intrinsically encourages the sparsity when removing the least likely candidate from a candidate labeling set.

\begin{figure}
\centering
\includegraphics[width=0.45 \textwidth]{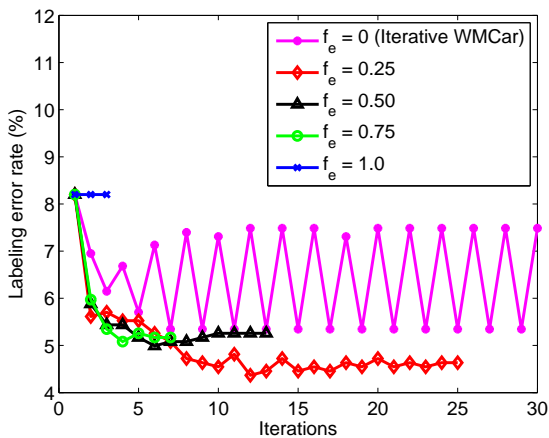}
\caption{Labeling error rate versus the number of iterations in WMCar-ICE. The performance is evaluated in the \emph{Lost} $(16,8)$ dataset with various elimination factors. The performance of WMCar-ICE ($f_e = 0$) fluctuates since the ICE procedure becomes ineffective as $f_e = 0$.}\label{fig:WMCar_ICE_thre}
\end{figure}

We conduct the sensitivity analysis of WMCar-ICE with $\lambda = \lambda_o$ and $\gamma = 2\lambda_o$ and evaluate the performance with various $f_e$.
In Figure \ref{fig:WMCar_ICE_thre}, the performance of WMCar-ICE ($f_e = 0$) fluctuates since the ICE procedure becomes ineffective as $f_e = 0$. A small elimination factor ($f_e = 0.25$) yields better performance than large elimination factors, but it takes more iterations to converge. Since the candidate elimination step in WMCar-ICE can incur an irreversible error, a small elimination factor can conservatively eliminate the least likely candidates in the candidate labeling sets. Hence, abrupt decision resulting from large elimination factors can be avoided, and the soft labeling matrix can be gently updated to guide the low-rank approximation of heterogeneous matrix. Figure \ref{fig:WMCar_ICE_thre} confirms that the selected parameters ($f_e =0.5$ and the maximum number of iterations equal to 5) yield good performance in terms of the rate of convergence and labeling error rate.
  
Owing to the greedy nature of the ICE procedure, we cannot guarantee that labeling error rates will monotonically decrease as the number of iterations of ICE increases. An incorrect removal of the actual label in the candidate label set leads to irreversible labeling error of this instance. Moreover, it can cause a detrimental effect on resolving the ambiguity of other instances since the error of an incorrectly labeled instance can be propagated to other instances of similar appearance through low-rank approximation of the heterogeneous matrix.

\subsection{Convergence}
Since the projection method in Section \ref{subsec:y_step} is not non-expansive, we cannot simply follow the rationale that the composition of gradient, shrinkage, and projection steps is non-expansive to prove convergence \cite{Cabral2014}. We attempt to replace the projection method of MCar with the Euclidean projection onto the simplex \cite{Duchi2008epo}, which is a non-expansive projection, but the performance of the modified MCar degrades significantly. An explanation is that the Euclidean projection onto the simplex can inadvertently generate non-sparse entries, which conflicts with the original objective to encourage the sparsity of the soft labeling matrix in (\ref{eqn:Y_relaxed1_W2}).
On the other hand, our simple projection step normalizes the $\ell_1$ norm of a soft labeling vector, which effectively restricts the soft labeling vector to lying on the $\ell_1$ ball and maintains an identical sparsity. Although the convergence of MCar has been observed empirically in \cite{Chen2015mcf}, a theoretical justification of convergence needs further investigation. Since the number of ambiguous labels is finite, the convergence of ICE is straightforward with $f_e > 0$

\section{Conclusions}\label{Chapter_MCar:sec:Con}
We introduced a novel matrix completion framework for resolving the ambiguity of labels. In contrast to existing iterative alternating approaches, the proposed MCar method ensures all the instances and their associated ambiguous labels are utilized as a whole for resolving the ambiguity. Since MCar is capable of discovering the underlying low-rank structure of subjects, it is robust to within-subject variations. Hence, MCar can serve as the counterpart of discriminative ambiguous learning methods. Besides, WMCar generalizes MCar to compensate for labeling imbalance, and thus an instance associated with minority labels has a stronger impact than that associated with majority labels.
The ICE procedure improves the performance of iterative WMCar by eliminating a portion of the least likely candidates in each iteration.
As demonstrated by the experiments on the synthesized ambiguous labels and two datasets collected from real world, our proposed methods consistently resolve the ambiguity when single face images or group of face images are ambiguously labeled.

\section*{Acknowledgments}
This research is based upon work supported by the Office of the Director of National Intelligence (ODNI), Intelligence Advanced Research Projects
Activity (IARPA), via IARPA R\&D Contract No. 2014-14071600012. The views and conclusions contained herein are those of the authors and should
not be interpreted as necessarily representing the official policies or endorsements, either expressed or implied, of the ODNI, IARPA, or the U.S. Government. The U.S. Government is authorized to reproduce and distribute reprints for Governmental purposes notwithstanding any copyright annotation
thereon.

\IEEEdisplaynontitleabstractindextext

% \fi
\IEEEpeerreviewmaketitle

\ifCLASSOPTIONcaptionsoff
  \newpage
\fi



%\bibliographystyle{IEEEtran}

%\bibliographystyle{ieee}
\bibliographystyle{IEEEtran}
\bibliography{MCJ}

% \end{thebibliography}


\vspace{-20pt}
\begin{IEEEbiography}[{\includegraphics[width=1in,height=1.25in,clip,keepaspectratio]{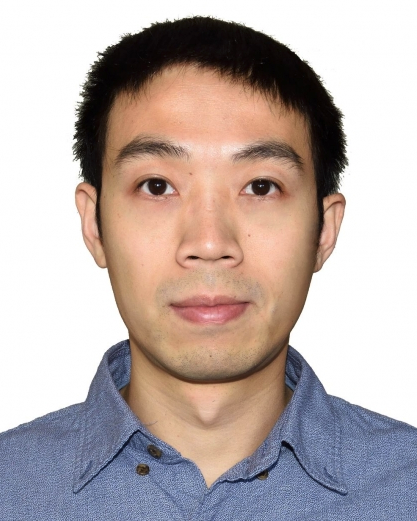}}]{Ching-Hui Chen}
received the B.S. degree in Electrical
Engineering and Computer Science Undergraduate
Honors Program and M.S. degree in Electronics
Engineering both from National Chiao Tung University,
Hsinchu, Taiwan. He received the Ph.D. degree in Electrical and Computer
Engineering from the University of Maryland,
College Park, MD, USA. His research interests are
in computer vision and pattern recognition, with
primary focuses on face recognition, stereo vision, and camera networks.
\end{IEEEbiography}
\vspace{-20pt}
\begin{IEEEbiography}[{\includegraphics[width=1in,height=1.25in,clip,keepaspectratio]{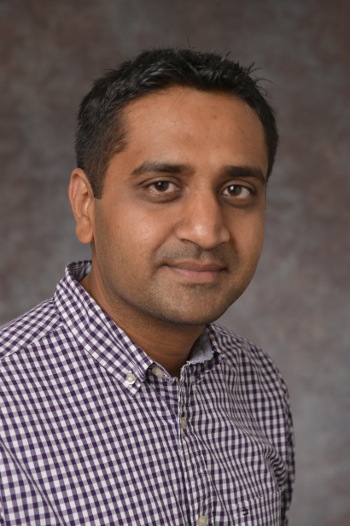}}]{Vishal M. Patel}
received the B.S. degrees in electrical engineering and applied mathematics (Hons.) and the M.S. degree in applied mathematics from North Carolina State University, Raleigh, NC, USA, in 2004 and 2005, respectively, and the Ph.D. degree in electrical engineering from the University of Maryland College Park, MD, USA, in 2010. He is currently an A. Walter Tyson Assistant Professor in the Department of Electrical and Computer Engineering (ECE) at Rutgers University.  Prior to joining Rutgers University, he was a member of the research faculty at the University of Maryland Institute for Advanced Computer Studies (UMIACS). His current research interests include signal processing, computer vision, and pattern recognition with applications in biometrics and imaging. He has received a number of awards including the 2016 ONR Young Investigator Award, the 2016 Jimmy Lin Award for Invention,  A. Walter Tyson Assistant Professorship Award, the Best Paper Award at IEEE BTAS 2015, and Best Poster Awards at BTAS 2015 and 2016.  He is an Associate Editor of the IEEE Signal Processing Magazine,  IEEE Biometrics Compendium, and serves on the Information Forensics and Security Technical Committee of the IEEE Signal Processing Society.   He is a member of Eta Kappa Nu, Pi Mu Epsilon, and Phi Beta Kappa.\end{IEEEbiography}
\vspace{-20pt}
\begin{IEEEbiography}[{\includegraphics[width=1in,height=1.25in,clip,keepaspectratio]{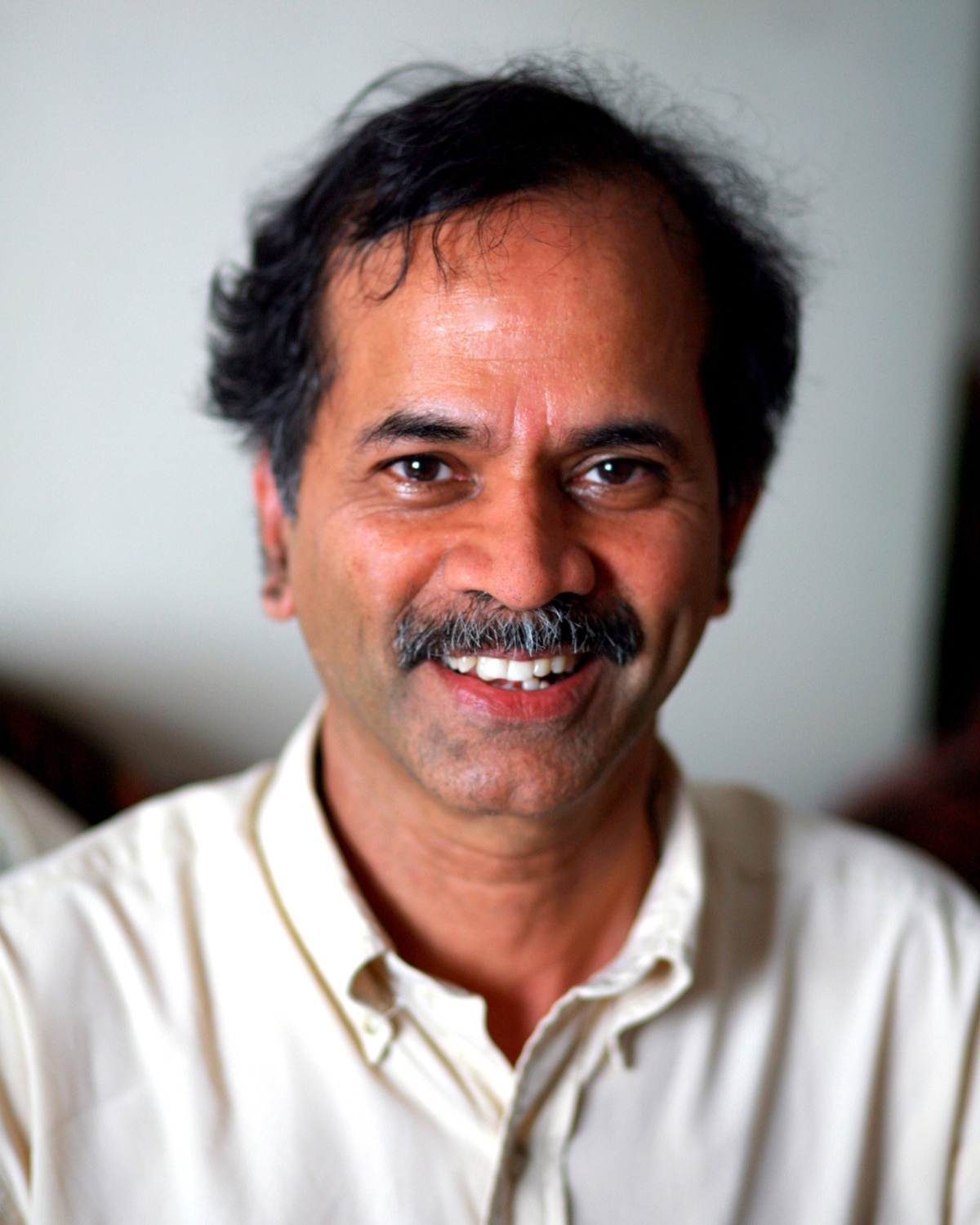}}]{Rama Chellappa}
is a Minta Martin Professor of Engineering and Chair of the ECE department at the University of Maryland. Prof. Chellappa received the K.S. Fu Prize from the International Association of Pattern Recognition (IAPR). He is a recipient of the Society, Technical Achievement and Meritorious Service Awards from the IEEE Signal Processing Society and four IBM faculty Development Awards. He also received the Technical Achievement and Meritorious Service Awards from the IEEE Computer Society. At UMD, he received college and university level recognitions for research, teaching, innovation and mentoring of undergraduate students. In 2010, he was recognized as an Outstanding ECE by Purdue University. Prof. Chellappa served as the Editor-in-Chief of PAMI. He is a Golden Core Member of the IEEE Computer Society, served as a Distinguished Lecturer of the IEEE Signal Processing Society and as the President of IEEE Biometrics Council. He is a Fellow of IEEE, IAPR, OSA, AAAS, ACM and AAAI and holds four patents.
\end{IEEEbiography}

%\newpage



%\vfill


\end{document}

%% file: short.tex
\newfont{\Bd}{msbm10 at 12 truept}
\newfont{\Sc}{eusm10 at 12 truept}

\def\diag{\mathop{\rm diag}\nolimits}